\documentclass[10pt,twocolumn,letterpaper]{article}

\usepackage[pagenumbers]{iccv} % To force page numbers, e.g. for an arXiv version
\usepackage{times}
\usepackage{epsfig}
\usepackage{graphicx}
\usepackage{amsmath}
\usepackage{amssymb}
\usepackage{makecell}
\usepackage{multirow}
\usepackage{booktabs} 
\usepackage{soul}
\usepackage{xcolor}
\usepackage{colortbl}
\definecolor{iccvblue}{rgb}{0.21,0.49,0.74}
\usepackage[pagebackref,breaklinks,colorlinks,allcolors=iccvblue]{hyperref}

\def\paperID{5526} % *** Enter the Paper ID here
\def\confName{ICCV}
\def\confYear{2025}

\newcommand{\ourpipeline}{{\textsc {LiFT}}\xspace}
\newcommand{\ourpipelinebold}{{\textbf{\textsc {LiFT}}}\xspace}
\newcommand{\ourmethod}{{\textsc {LiFT-Critic}}\xspace}
\newcommand{\ourmethodbold}{{\textbf{\textsc {LiFT-Critic}}}\xspace}
\newcommand{\ourdataset}{{\textsc {LiFT-HRA}}\xspace}
\newcommand{\ourdatasetbold}{{\textbf{\textsc {LiFT-HRA}}}\xspace}
\newcommand{\ourdatasetit}{{\textit{LiFT-HRA}}\xspace}
\definecolor{Apricot}{rgb}{1.0, 0.85, 0.6}

\newcommand\mypara[1]{\vspace{1mm}\noindent\textbf{#1}}
\newcommand{\app}[1]{{\color{purple} #1}\xspace}

\def\ie{\emph{i.e., }}
\def\eg{\emph{e.g., }}

\title{LiFT: Leveraging Human Feedback for Text-to-Video Model Alignment}
\author{
Yibin Wang $^{1,2}$ \quad
Zhiyu Tan $^{1,2}$ \quad
Junyan Wang $^{3}$ \quad 
Xiaomeng Yang $^{2}$ \\ [2pt]\quad 
Cheng Jin $^{1}$\footnotemark[2]  \quad
Hao Li $^{1,2}$\footnotemark[2] \\ [6pt]
{
$^{1}$ {Fudan University} \quad $^{2}$ {Shanghai Academy of Artificial Intelligence for Science}
}  \\
{
$^{3}$ {Australian Institute for Machine Learning, The University of Adelaide} 
}
\\
\href{https://codegoat24.github.io/LiFT/}{codegoat24.github.io/LiFT}
} 

\begin{document}
\maketitle

\begin{abstract}
Recent advances in text-to-video (T2V) generative models have shown impressive capabilities. 
However, these models are still inadequate in aligning synthesized videos with human preferences (\eg accurately reflecting text descriptions), which is particularly difficult to address, as human preferences are subjective and challenging to formalize as objective functions. Existing studies train video quality assessment models that rely on human-annotated ratings for video evaluation but overlook the reasoning behind evaluations, limiting their ability to capture nuanced human criteria. Moreover, aligning T2V model using video-based human feedback remains unexplored.
Therefore, this paper proposes \ourpipelinebold, the first method designed to leverage human feedback for T2V model alignment. Specifically, we first construct a Human Rating Annotation dataset, \ourdatasetbold, consisting of approximately 10k human annotations, each including a score and its corresponding reason.
Based on this, we train a reward model \ourmethodbold to learn reward function effectively, which serves as a proxy for human judgment, measuring the alignment between given videos and human expectations.
Lastly, we leverage the learned reward function to align the T2V model by maximizing the reward-weighted likelihood. 
As a case study, we fine-tune CogVideoX-2B using our pipeline, surpassing CogVideoX-5B across all 16 metrics, demonstrating the effectiveness of human feedback in enhancing video quality.

\end{abstract}

\section{Introduction}
Recent advancements in Text-to-Video (T2V) generation models \cite{wang2024videocomposer,gupta2025photorealistic,wang2023lavie,wang2023modelscope,yang2024cogvideox,he2024venhancer,zhou2024allegro} have achieved remarkable results. 
These models enable users to generate high-quality videos, offering a flexible, controllable approach to video creation. 
% text感觉描述不清，因为下一句话提出了text description的问题
Despite this progress, these models still face challenges such as artifacts, misalignment with semantic requirements, and unnatural motion \cite{Li2024T2VTurboBT,he2024videoscore}. These issues primarily arise from the inherent subjectivity of human preferences, which are difficult to formalize as objective functions \cite{Kou2024SubjectiveAlignedDA}. Consequently, effectively incorporating human preferences into T2V models remains a major challenge.
\begin{figure}[t]

    \centering
    \includegraphics[width=0.85\linewidth]{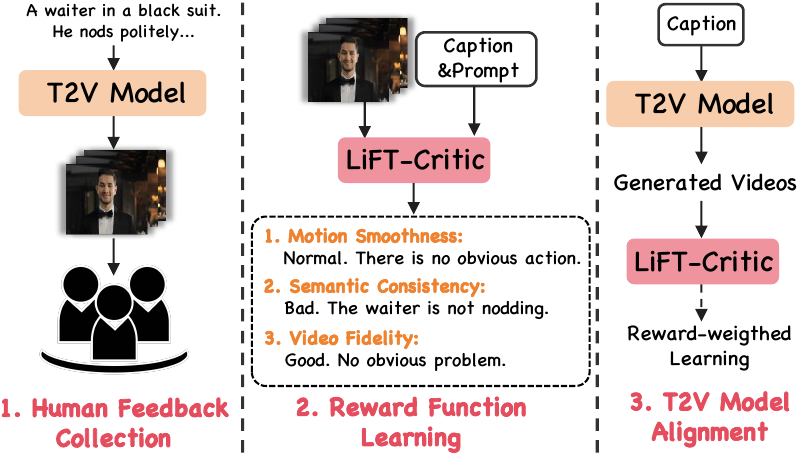}
    \caption{\textbf{An illustration of the proposed \ourpipeline.} First, we construct a comprehensive human feedback dataset. Then, a reward model is trained to learn the reward function. Finally, the T2V model is fine-tuned by the reward model to align its output with human expectations.
    % The overview of \ourpipeline. First, we collect human feedback by generating video-text pairs using prompts, followed by human annotations to construct a comprehensive feedback dataset. Next, a reward model is trained on this dataset to predict human preferences. Finally, the T2V model is fine-tuned using the learned reward function to align its output with human expectations.
    }
    \label{fig:intro}
    \vspace{-0.5cm}
\end{figure}

% 还是，text的问题是怎么联系到human preferences了？我有点没看懂...

In language modeling and the text-to-image domain, learning from human feedback to align model outputs \cite{lee2023aligning,liang2024rich,bai2022training,ouyang2022training} has demonstrated efficiency. Building on this success, recent efforts in T2V generation \cite{yuan2024instructvideo,Li2024T2VTurboBT} have incorporated feedback from image reward models \cite{wu2023human,radford2021learning} into the training process. However, image reward models fail to capture crucial temporal aspects of video, such as motion continuity and smooth transitions between frames, thereby struggling to align T2V models with human expectations. To this end, methods like T2VQA \cite{Kou2024SubjectiveAlignedDA} and VideoScore \cite{he2024videoscore} train video quality assessment models based on human-annotated video ratings to better evaluate video's visual quality, temporal consistency, and so on by assigning a score to each dimension.
However, these studies face two main challenges: 
(1) \textbf{Lack of interpretability}: by focusing solely on assessment results (only learning ratings), these models overlook the underlying reasoning, hindering their ability to capture nuanced human criteria.
(2) \textbf{Limited human feedback guidance}: Utilizing video-based human feedback for aligning T2V models remains an unexplored area, restricting their ability to meet complex and diverse human expectations.

We argue that the key to aligning T2V models with human expectations lies in constructing an effective pipeline that integrates human feedback into the training process. This pipeline should consist of two key components: a high-quality dataset where each annotation not only provides ratings for synthesized videos but also explains the reason, and a reward model that learns a human feedback-based reward function. The dataset is essential for capturing nuanced human evaluation criteria beyond simple numerical scores, while the reward model ensures accurate interpretation of both explicit ratings and underlying human judgments rather than relying on superficial correlations. With this foundation, the reward model can then be leveraged to fine-tune T2V models, providing a reliable optimization signal that guides them toward generating videos that better align with complex and diverse human preferences.

% These challenges are interconnected: the lack of interpretability in existing reward models prevents them from reliably learning a reward function that truly reflects human preferences, thereby limiting the effective alignment of T2V models with human expectations. 
% We posit that \jy{the core issue} lies in the absence of human feedback-based datasets where each annotation for synthesized videos includes both ratings and their underlying reasoning. Such datasets are critical for training reward models not only to understand the score but also the reason behind human judgments. Without this explanatory context, models are limited to surface-level evaluations, focusing solely on the final score, thereby failing to grasp the deeper, often subjective criteria that influence human judgments.
% By developing such a dataset, a reliable reward model could be trained, which could then be used to fine-tune T2V models for effective alignment with human expectations.
% 这一段我有疑问，core issue咱们应该说只有dataset吗？然后最后一句话点出reward model感觉有点弱？\ourdatasetbold和\ourmethodbold还有 alignment平等的说是不是好一点？

Therefore, this paper takes the first step to propose a novel fine-tuning method, \ourpipelinebold, leveraging human feedback for T2V model alignment through three key stages, as illustrated in Fig. \ref{fig:intro}: (1) \textbf{Human Feedback Collection}: generate video-text pairs through diverse prompts, followed by detailed human annotations to build a comprehensive feedback dataset; (2) \textbf{Reward Function Learning}: train a reward model capturing human preferences based on this dataset to predict human feedback scores; and (3) \textbf{T2V model alignment}: fine-tune the T2V model using the learned reward function to optimize its output in alignment with human expectations. Specifically, we first categorize human preferences into three key dimensions: semantic consistency, motion smoothness, and video fidelity. For these dimensions, we propose \ourdatasetbold dataset that collects 10K human feedback annotations, each including both ratings and the corresponding reasoning. We then train a reward model, \ourmethodbold, based on the large multimedia model (LMM) \cite{lin2024vila} to learn human feedback-based reward function. Unlike existing assessment models \cite{he2024videoscore,Kou2024SubjectiveAlignedDA}, our model not only learns to predict ratings but also captures the reason behind them, thereby improving interpretability and providing a deeper understanding of the evaluation process. Finally, we apply this learned reward function to evaluate the quality of T2V model outputs and update the model using reward-weighted likelihood maximization.

In summary, this work introduces \ourpipeline, the first fine-tuning pipeline for aligning T2V models with human feedback. To achieve this, we construct \ourdataset, a comprehensive dataset where each annotation provides both ratings and detailed reason, enabling models to capture nuanced human evaluation criteria. Additionally, we develop \ourmethod, a reward model designed to accurately learn a human feedback-based reward function, effectively bridging the gap between subjective human preferences and model optimization. Leveraging these components, we apply our pipeline to fine-tune CogVideoX-2B \cite{yang2024cogvideox}, demonstrating that the resulting model surpasses the larger CogVideoX-5B across all 16 metrics of the VBench \cite{Huang2023VBenchCB} benchmark, significantly improving alignment with human preferences.

\begin{figure*}[!thb]

    \centering
    \includegraphics[width=0.95\linewidth]{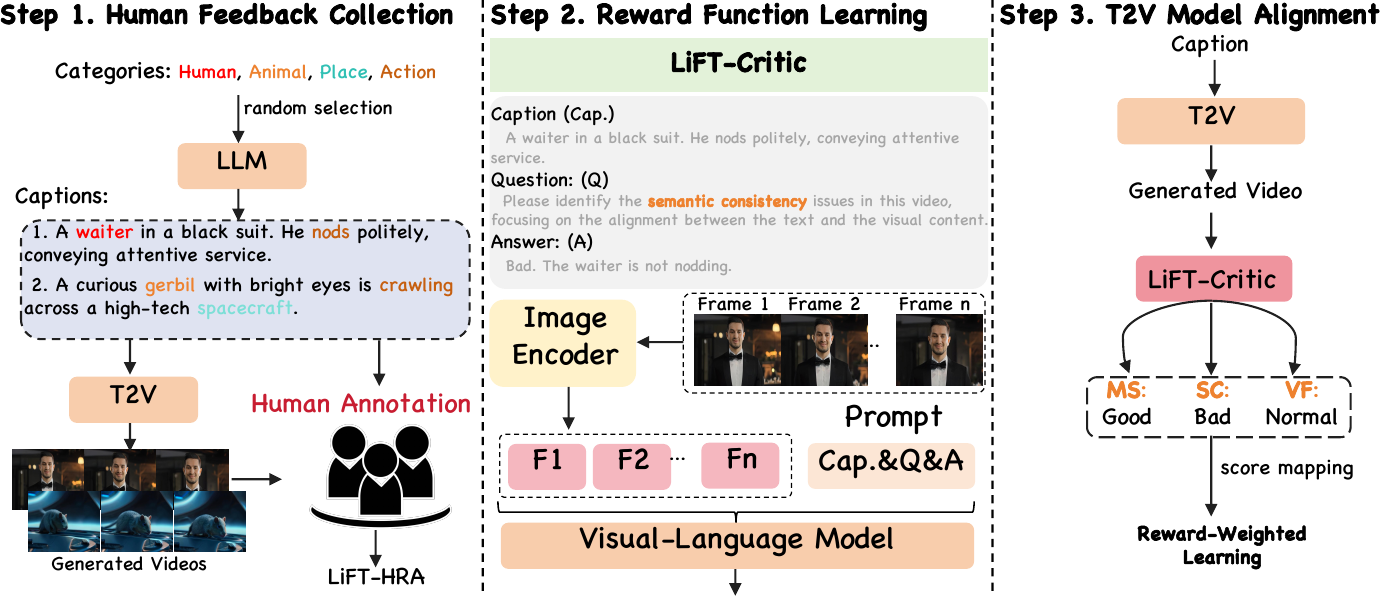}
    \caption{\textbf{The overview of our proposed pipeline}. This illustration depicts three key steps of our fine-tuning pipeline: (1) Human Feedback Collection: we generate video-text pairs using prompts expanded from random category words with an LLM, then annotate them to create \ourdataset. (2) Reward Function Learning: a visual-language model \ourmethod, is trained to predict human preference scores across three dimensions, learning the reward function from the dataset. (3) T2V Model Alignment: \ourmethod evaluates the T2V-generated videos, assigns scores, and maps them into a reward weight to fine-tune the T2V model, aligning it with human preferences.
% (1) Human Feedback Collection. We start by selecting phrases derived from randomly chosen category words and expanding them into detailed prompts using an LLM. These prompts are then used by a T2V model to generate video-text pairs, which humans subsequently annotate to construct \ourdataset.
% (2) Reward Function Learning. Based on this dataset, we train a Visual-Language model, \ourmethod, to predict scores across three dimensions, effectively learning a reward function that reflects human preferences.
% (3) T2V Model Alignment. Finally, \ourmethod assesses the videos generated by the T2V model, assigning scores across the defined dimensions. These scores are then mapped into a reward weight, which guides the fine-tuning of the T2V model through reward-weighted learning, enabling it to better align with human preferences.
}

    \label{fig:pipeline}
    \vspace{-0.5cm}
\end{figure*}

\section{Related Work}

\mypara{Diffusion-based T2V models}.
% Text-to-video (T2V) generation aims to synthesize high-quality videos conditioned on text descriptions. 
Recent advancements in T2V focus on leveraging large-scale datasets to train more robust models \cite{wang2024videocomposer,guo2023animatediff,chen2023gentron,gupta2025photorealistic,wang2023lavie,wang2023modelscope}. A common approach involves adapting pre-trained text-to-image (T2I) models by introducing temporal layers and fine-tuning them on video datasets or employing a joint image-video training strategy \cite{blattmann2023align,wang2023modelscope,wang2023lavie}.
% To enhance video quality, multi-stage inference pipelines are often utilized \cite{ho2022imagen,wang2023lavie,blattmann2023align}, including cascaded video diffusion models. These typically consist of a T2V base model followed by frame interpolation and video super-resolution models \cite{blattmann2023align,wang2023lavie,zhou2024upscale}. 111
Despite their effectiveness, such pipelines are limited in improving spatiotemporal resolution. To address this, recent works \cite{yang2024cogvideox,zhou2024allegro,chen2023videocrafter1,he2024venhancer} offer diverse solutions. 
For example, VEnhancer \cite{he2024venhancer} unifies temporal and spatial super-resolution with refinement in a single model.
% Meanwhile, CogVideoX \cite{yang2024cogvideox} introduces a scalable 3D causal VAE combined with an expert transformer to generate coherent, action-rich, and long-duration videos. These advancements collectively push the boundaries of video generation in terms of quality, coherence, and scalability.
However, these models still remain inadequate in aligning synthesized videos with human preferences, such as accurately reflecting text descriptions. We thus train a reward model to capture human preferences and then use it to fine-tune T2V models. Related studies will be introduced in the following.

\mypara{Vision-and-language reward models}. 
In T2I generation, several studies aim to incorporate human preferences into model evaluation and optimization \cite{kirstain2023pick, xu2024imagereward, li2023agiqa, wu2023human, zhang2024learning}.
For instance, \cite{kirstain2023pick,xu2024imagereward} collects user-generated prompts and preference annotations via a web platform. 
% \cite{li2023agiqa,wu2023human,zhang2024learning} evaluates overall and alignment preferences for diffusion, GAN, and auto-regressive models.
% For the T2V generation, researchers have worked on evaluation methods to benchmark the generative models \cite{Huang2023VBenchCB,liu2024evalcrafter,Bansal2024VideoPhyEP}.
For T2V generation, recent works \cite{yuan2024instructvideo,Li2024T2VTurboBT} integrate feedback from image reward models \cite{wu2023human,radford2021learning} into the training process. However, image-based human feedback overlooks crucial temporal aspects of video which are vital for generating coherent and natural videos. To bridge this gap, methods such as T2VQA \cite{Kou2024SubjectiveAlignedDA} and VideoScore \cite{he2024videoscore} train video quality assessment models directly on human-annotated video ratings. 
However, these models focus solely on assessment results, overlooking the interpretability of the evaluation process. 
% This limits their ability to capture nuanced criteria and effectively align with human preferences. 
Thus, we propose a novel reward model that evaluates synthesized videos by providing both ratings and corresponding reasons.

\mypara{Aligning generative model use reward learning}. 
Leveraging reward models, various approaches have been developed to align visual generative models with human preferences. These methods include reinforcement learning-based techniques \cite{fan2024reinforcement,Zhang2024LargescaleRL,black2023training,chen2025enhancing} and reward fine-tuning approaches \cite{Clark2023DirectlyFD,Li2024RewardGL,Yuan2023InstructVideoIV,lee2023aligning,prabhudesai2024video}. By integrating feedback mechanisms, these methods aim to enhance model outputs to better reflect human preferences. 
To date, video-based human feedback has not been fully leveraged to align T2V models, limiting their ability to meet the complex and diverse expectations of users. Therefore, we explore a novel fine-tuning pipeline to align T2V models with human preferences.

% For instance, T2V-Turbo \cite{Li2024T2VTurboBT} combines existing reward models to simulate human feedback effectively, enabling the fine-tuning of models to align more closely with user preferences. Additionally, multi-objective reward models, such as those proposed by \cite{Wang2024ArithmeticCO,Wang2024InterpretablePV}, use regression heads to capture nuanced human preferences across multiple dimensions.
% Data distillation is another widely used strategy, where models are fine-tuned on high-quality datasets \cite{Li2024T2VTurboBT}. This ensures the generative outputs exhibit higher fidelity and alignment with desired characteristics. Recently, Diffusion-DPO \cite{wallace2024diffusion} extends Direct Preference Optimization to train diffusion models using preference-labeled data, offering a direct pathway to incorporate human feedback into the model's training process.

% ~\cite{lee2023aligning} ~\cite{fan2024reinforcement} ~\cite{black2023training} ~\cite{yang2024using} ~\cite{sun2023dreamsync} ~\cite{chen2025enhancing} ~\cite{yoon2024censored}

% ~\cite{yuan2024instructvideo}  ~\cite{prabhudesai2024video}  

% \subsection{Evaluating generated vidoes} 

% \newpage
\section{\ourpipelinebold}
\subsection{Overview}
This work aims to integrate human feedback into the training of text-to-video (T2V) models, a challenging task due to the inherent subjectivity of human preferences, which are difficult to formalize as objective functions. Previous research, such as \cite{wu2023human} and \cite{radford2021learning}, has explored the use of image-based reward models for aligning T2V models. However, these reward models are difficult to capture the crucial temporal dynamics of video, thereby limiting their ability to satisfy the complex expectations of video generation. To this end, this work proposes \ourpipeline, the first fine-tuning pipeline designed to align T2V models with human feedback. As illustrated in Fig. \ref{fig:pipeline}, our pipeline consists of three sequential stages: (1) Human Feedback Collection (Sec. \ref{sec: dataset}): We first collect human feedback and construct a comprehensive dataset \ourdataset , where each annotation provides both numerical ratings and detailed reasoning. (2) Reward Function Learning (Sec. \ref{sec:vidcritic}): We then train a reward model that learns a human feedback-based reward function based on this dataset to capture nuanced evaluation criteria. (3) T2V Model Alignment (Sec. \ref{sec:align}): Finally, the learned reward function is leveraged to fine-tune the T2V model, enabling it to better align with human expectations.

% \jy{Specifically, we first collect human feedback and construct \ourdataset (Sec. \ref{sec: dataset}), a comprehensive dataset where each annotation provides both numerical ratings and detailed reasoning. This dataset serves as the foundation for training \ourmethod, a reward model that learns a human feedback-based reward function to capture nuanced evaluation criteria (Sec. \ref{sec:vidcritic}). Finally, the learned reward function is leveraged to fine-tune the T2V model (Sec. \ref{sec:align}), enabling it to generate videos that better align with human expectations.}

% \subsection{Overview}
% \jy{This section introduces our fine-tuning pipeline to align text-to-video (T2V) models with human preferences using human feedback. The overall approach, illustrated in Fig. \ref{fig:pipeline}, comprises three key steps:
% First, we collect human feedback and curate the \ourdataset dataset (Sec. \ref{sec: dataset}). This dataset serves as the foundation for training \ourmethod, which learns a reward function that captures human preferences (Sec. \ref{sec:vidcritic}). Finally, we utilize the learned reward function to refine the T2V model (Sec. \ref{sec:align}), ensuring improved alignment with human evaluation criteria.}
% 我觉得可以写一小节problem statement或者overview把整个human feedback这个任务的pipline整体描述清楚，毕竟咱们是第一篇工作，强调的是first，然后再简单说一下每个section

\subsection{\ourdatasetbold: Human Feedback Collection} \label{sec: dataset}
Existing human feedback-based text-to-video datasets \cite{he2024videoscore, Kou2024SubjectiveAlignedDA} primarily focus on outcome evaluation, such as overall video quality or video-text alignment, by collecting single evaluation scores. While effective for training models to assess high-level metrics, these datasets lack critical insights into the evaluation process, \ie the reasoning behind the assigned scores. For reward function learning, we argue that incorporating such reasoning is crucial for accurately aligning with nuanced human preferences. To address this gap, we introduce \ourdatasetbold, a new dataset tailored for reward model training. This dataset combines evaluation scores with their corresponding reason, enabling a more holistic and interpretable alignment with human expections, which will be elaborated in the following.

\mypara{Video-Text Dataset}.
Our reward model is designed to evaluate synthesized videos based on human preferences. Therefore, constructing a comprehensive video-text dataset is essential. Specifically, we start by generating a set of diverse prompts. This involves creating selection lists for different categories such as humans, animals, places, simple actions, and complex actions. For each prompt, as shown in Fig. \ref{fig:pipeline} (left), we randomly choose 1–2 subjects from the human (\eg waiter) and animal (\eg gerbil) categories, a scene (\eg spacecraft) from the places list, and an action (\eg  nod) from either the simple actions or complex actions categories. These selected elements are combined into a phrase and refined into a detailed textual description using LLM \cite{yang2024qwen2}. Finally, multiple videos are generated for each prompt using T2V models, forming the video-text dataset.
% prompt描述不清晰，每个element没有突出，还有可以加个e.g., 我觉得这个挺重要的，甚至可以加个小图

% 这一段的内容其实就是那张图，我觉得用不了一个小节来介绍，可以和下一节data correction整合一下
% To collect comprehensive human evaluations, we categorize human preferences into three key dimensions: semantic consistency, motion smoothness, and video fidelity. For each video-text pair in the dataset, we enlist annotators to evaluate the generated videos across these dimensions. The annotation UI is illustrated in Fig. \ref{fig:annotation_ui}. Specifically, given videos and their captions, annotators assess each video by assigning a score (Good, Normal, or Bad) in each dimension based on predefined scoring criteria and provide a detailed reason for their evaluation.
\begin{figure}[t]

    \centering
    \includegraphics[width=0.9\linewidth]{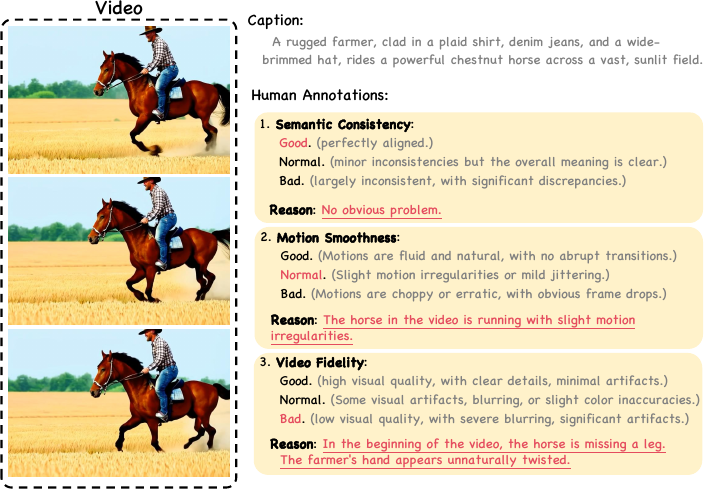}
    \caption{\textbf{An illustration of our annotation UI}. Annotators evaluate each video by assigning scores to each dimension and providing the reason behind their assessments.}
    \label{fig:annotation_ui}
    \vspace{-0.6cm}
\end{figure}

% \mypara{Human Annotation and Data Correction}.
\mypara{Annotation}.
To collect comprehensive human evaluations, we categorize human preferences into three key dimensions: semantic consistency, motion smoothness, and video fidelity. For each video-text pair in the dataset, we enlist annotators to evaluate the generated videos across these dimensions by assigning a score in each dimension and providing a detailed reason. The annotation UI is illustrated in Fig. \ref{fig:annotation_ui}. After all data has been annotated, we perform a three-stage correction process to ensure reliable data quality:
1) \textit{Coarse Filtering:} Annotators manually review each labeled sample, removing those with obvious annotation errors or misaligned feedback that does not adhere to the evaluation criteria, ensuring data quality.
2) \textit{Iterative Refinement:} 
The dataset is then split into two halves. One half trains an initial reward model, which is then used to annotate the other half. If the model's output aligns with human annotations, they are retained; otherwise, human annotators decide which to keep. The corrected data is used to retrain the model, refining the other half.
3) \textit{Final Integration:} Using the fully corrected dataset, we train a final reward model. This model is then used to re-annotate the data that was removed during the first stage, and the newly annotated data is incorporated back into the dataset. 
More details of \ourdataset construction are provided in Appendix \app{A}.

\begin{figure*}[t]

    \centering
    \includegraphics[width=1\linewidth]{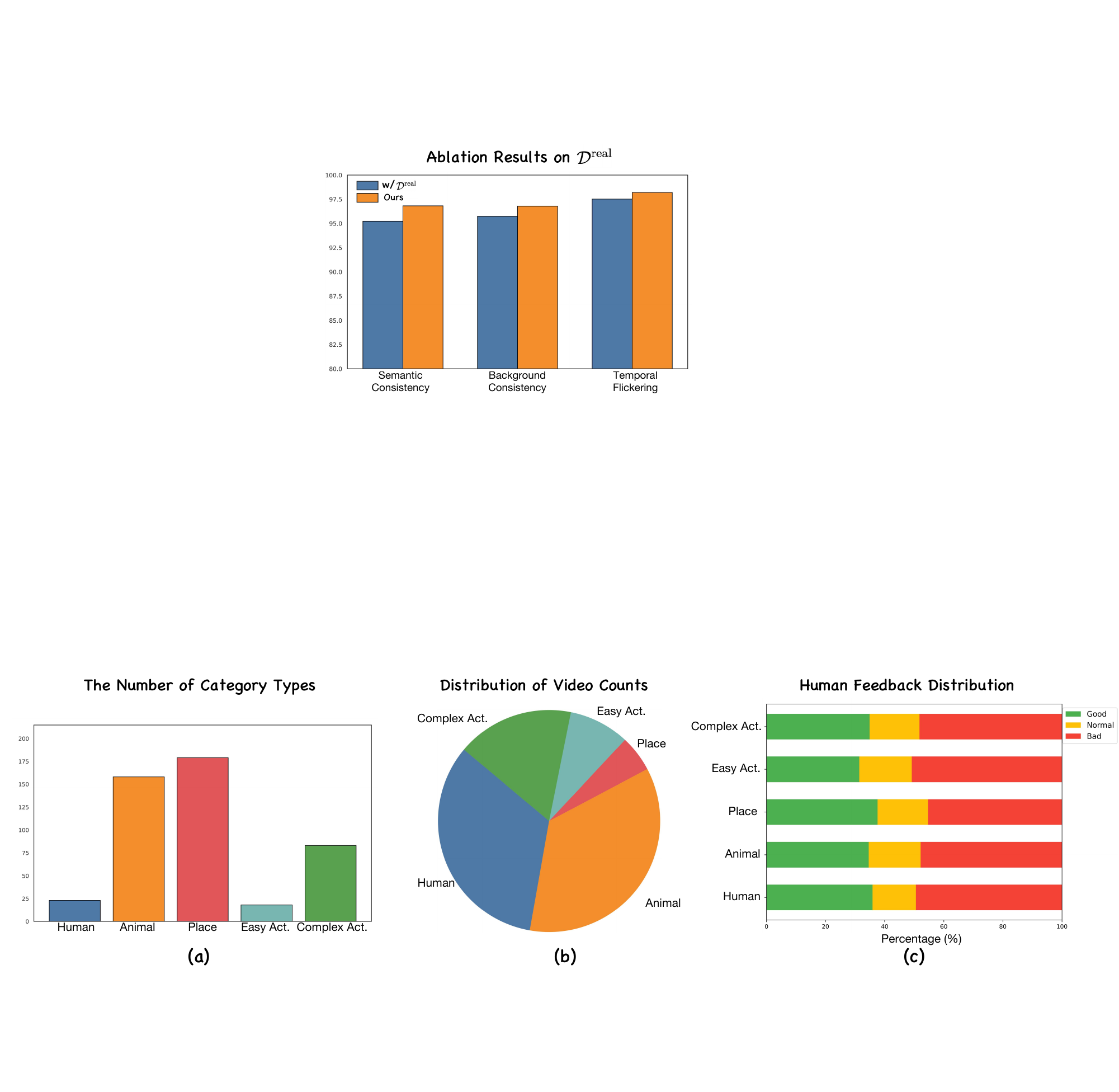}
    \caption{\textbf{The visualized statistic results of our proposed \ourdataset.} It illustrates the distribution of category types, the video count across these categories, and the corresponding human feedback distribution for each category.}
    \label{fig:dataset}
    \vspace{-0.5cm}
\end{figure*}

\mypara{Statistical Analysis}.
Finally, we collect 10K high-quality human feedback video quality question-answer pairs, forming the \ourdatasetbold dataset. Its statistical analysis is visualized in Fig. \ref{fig:dataset}. Specifically, (a) the dataset covers a diverse range of categories, (b) each category contains multiple video types with varying numbers of samples, and (c) human feedback is distributed across all categories, with the majority of videos rated as ``Bad'', followed by ``Good'' and ``Normal'' in descending order. This indicates that, on average, the videos evaluated tend to have more room for improvement in terms of their alignment with human preferences. Additionally, the variation in feedback distribution across categories highlights different challenges and complexity in evaluating videos from diverse domains, such as actions, animals, and places. This variability is crucial for training robust reward models that can generalize across different video types and better align with human expectations.

Overall, the diversity and comprehensiveness of \ourdataset provide a robust foundation for accurately capturing human preferences across diverse contexts.

% This dataset goes beyond simple scoring by incorporating detailed human reasoning, enabling the reward model to grasp nuanced evaluation criteria and better align with human preferences. 
% The process of training the reward function based on this dataset is elaborated below.

% \subsubsection{Annotation Pipeline}

\subsection{\ourmethodbold: Reward Function Learning} \label{sec:vidcritic}
% To align with human preferences, we propose VidCritic, which utilizes the large-scale visual-language model (VLM) \cite{lin2024vila} with strong video understanding capabilities to learn the reward function from our curated dataset. This approach treats evaluation as a discriminative skill tailored for video quality assessment.
% Unlike existing methods that modify model architectures, such as adding prediction heads \cite{he2024videoscore,liang2024rich}, we retain the original structure of the multimedia model during fine-tuning, since architectural modifications can constrain the model’s inherent strengths and require larger datasets and longer training times. By preserving the original architecture, our method efficiently aligns the model with human feedback while maintaining its strong video-understanding capabilities.
% To construct a more effective reward model for aligning video generation with human feedback, we fine-tune a pre-trained Visual-Language Model (VLM)\cite{lin2024vila}  using annotated data. The reward model evaluates generated videos based on predefined dimensions by assigning scores and providing reason that reflect human preferences.
Based on the constructed high-quality dataset, we fine-tune a pre-trained Visual-Language Model (VLM) \cite{lin2024vila} to develop our reward model, \ourmethod, that learns a reward function based on human feedback. 
We transform scores into natural language labels (\eg ``Good'', ``Normal'', ``Bad'') and predict these labels through text generation, aiming to not only predict scores but also provide explanations. Current studies \cite{he2024videoscore,liang2024rich} typically add a regression head to VLM for direct score prediction. While straightforward, it fails to capture the complexity of human judgment, which involves nuanced, subjective criteria.
% \jy{For this purpose, two main approaches can be employed: (1) adding a regression head to the VLM for direct score prediction or (2) transforming scores into natural language labels (e.g., ``good'', ``normal'', ``bad'') and predicting these labels through text generation. This work adopts the second approach, aiming to bridge the gap by not only predicting scores but also providing explanations for the ratings.
% Current studies \cite{he2024videoscore,liang2024rich} typically follow the first approach, focusing on direct score prediction without addressing the underlying reasoning. While this method is straightforward, it fails to capture the complexity of human judgment, which often involves nuanced, subjective criteria.}
% In contrast, we adopt the second method, aiming to bridge this gap by training the model not only to predict scores but also to provide the reason behind them. 
% By transforming the scores into natural language labels, our model not only assigns ratings but also offers explanations for those ratings. 
In contrast, our approach improves the interpretability of the reward model by capturing the underlying reasoning behind human evaluations. Additionally, the text generation method aligns naturally with the pre-trained VLM's strengths in multimodal understanding and natural language generation, allowing the model to leverage its capabilities in producing semantically rich, contextually meaningful outputs.
% These explanations help the model understand the underlying criteria and human preferences, enabling it to better capture the fine-grained aspects of video evaluation. 

% Among these, the second approach demonstrates clear advantages in building a more robust and effective reward model. \cite{he2024videoscore,liang2024rich}
% The text generation method aligns naturally with the pre-trained VLM’s strengths in multimodal understanding and natural language generation. 
% This approach fully leverages the VLM’s ability to produce semantically rich and contextually meaningful outputs, enabling the reward model to better capture human preferences.
% By converting numerical scores into discrete labels, the method also reduces the sensitivity of the reward model to noisy annotations, ensuring stable and consistent performance.
% Furthermore, the text generation approach allows the reward model to incorporate nuanced context from input data, producing outputs that align more closely with human judgment. This capability enhances the reward model’s ability to generalize across varied scenarios and datasets, providing more reliable guidance for video generation models. By adopting this method, the reward model achieves a stronger alignment with human feedback, ultimately enabling more effective optimization of video generation quality.

Fig. \ref{fig:pipeline} (Step 2) illustrates our supervised fine-tuning process. The training begins by passing the video frames through an image encoder to extract image tokens. These image tokens, along with the textual caption, question, and answer, are then input into the pre-trained Visual-Language Model (VLM). Formally, the training sample consists of a triplet: \textbf{multimodal input} $M$, which includes both the video and its textual description; \textbf{question} $Q$; and \textbf{answer} $A$.
The optimization objective is based on the autoregressive loss function commonly used in training large language models, but it is applied exclusively to the answer.
% 这部分我感觉才是这节的重点，但是我觉得描述的太少了，应该根据图2一步步的详细的讲解，必要的时候可以添加各种在problem statement（overview）中的notation
The loss function can be expressed as:
\begin{align}
    L(\phi) = -\sum^N_{i=1} \log p(A_i \mid Q, M, A_{<i}; \phi),
\end{align}
where $N$ is the length of the ground-truth answer.
% , $Q$ is the question derived from the generated video and its available answer, $M$ includes the video and textual description, and $A$ is the human-annotated answer selected from the provided options.
To enhance the model's performance, this work enriches each answer option with detailed reasoning, training the model to predict multi-level scores and generate justifications for its judgments. This added context helps the VLM better understand the meaning of each option, thereby improving its alignment with human preferences. 

% 我不建议结果论的证明放在method中

% During training, VidCritic processes an evaluation prompt composed of the synthesized video, its caption, the evaluation question, and the corresponding model response. The training sample is organized as a sequence: 
% \[
% \textit{(Synthesized Video, Caption, Question, Score, Reason)}.
% \]
% A standard cross-entropy loss is applied to both the predicted scores and the corresponding justifications, ensuring that the model learns to provide accurate and interpretable evaluations.

After training, the reward model can predict a reward score for a given synthesized video and its corresponding caption. Specifically, \ourmethod evaluates the video across three dimensions, including \textit{semantic consistency}, \textit{motion smoothness}, and \textit{video fidelity}. Since these scores are multi-level qualitative evaluations (\eg \textit{Good}, \textit{Normal}, \textit{Bad}), a reward score mapping function \(s\), is employed to translate these qualitative assessments into numerical values. 
% 这两句话就很突兀了，\ourmethod在全文中第一次出现，就能can和evaluate。我没看懂为什么。
Formally, the reward score for a specific dimension \(d \in D\) is computed as:  $s(r_{\phi, d}(\mathbf{\textit{\textbf{x}}}, z))$,
where \(r_\phi\) represents the reward model with parameters $\phi$.
Finally, the overall reward score for a synthesized video is obtained by averaging the scores across all dimensions:  
\begin{equation} \label{equa:score}
r_\phi(\mathbf{\textit{\textbf{x}}}, z) = \frac{1}{|D|} \sum_{d \in D} s(r_{\phi, d}(\mathbf{\textit{\textbf{x}}}, z)),
\end{equation}  
where \(\mathbf{\textit{\textbf{x}}}\) is the synthesized video, and \(z\) is the associated caption.
The obtained reward score can effectively substitute human evaluations, providing feedback for the alignment of T2V models with human expectations.
% 这句话我觉得有点没必要，可以缩略成一句简短的话。然后主要写一下这部分的作用，可以把上面一些优点放到最后来写，前面主要还是讲解下方法过程

\subsection{T2V Model Alignment} \label{sec:align}
While the text-to-image domain has showcased the potential of human feedback-based alignment \cite{lee2023aligning}, extending this paradigm to video generation poses unique challenges due to the increased dimensionality and the intricate temporal dynamics inherent in video data.
Therefore, this work explores novel methods that employ our video reward model for T2V model alignment:

\noindent\textbf{Reward-Weighted Learning (RWL)}.
% therefore之后的内容感觉跟前面的内容不搭嘎，我建议丰富一下前面的，然后清晰一些后面的
% integrating two complementary training strategies: Reward-Weighted Learning (RWL) and Reject Sampling (RS). 
The core idea of RWL is to use the reward model $r_\phi$ to guide the learning of the T2V model $p$ with parameters $\theta$ by adjusting the likelihood of generated outputs. Specifically, the T2V model is trained to maximize the likelihood of generating videos that align with higher human-provided rewards, which reflect the more accurate human preferences. Inspired by \cite{lee2023aligning}, this process can be formulated as follows:
% 我记得咱们之前聊过这个，把这部分再细化一下，可以inspired，怎么引出的这个公式
\begin{equation}
\begin{aligned}
\mathcal{L}(\theta) = &\ 
\mathbb{E}_{(\textbf{\textit{x}}, z) \sim \mathcal{D}^{\mathrm{syn}}}
\left[-r_\phi(\textbf{\textit{x}}, z) \log p_\theta(\textbf{\textit{x}}|z)\right] \\
&+ 
\lambda \cdot \mathbb{E}_{(\textbf{\textit{x}}, z) \sim \mathcal{D}^{\mathrm{real}}}
\left[-\log p_\theta(\textbf{\textit{x}}|z)\right],
\end{aligned}
\end{equation}
where \(r_\phi(\textbf{\textit{x}}, z)\) computed by
Eq. \ref{equa:score} denotes the reward score for synthesized video \textbf{\textit{x}} with caption \textit{z}.
% the score function \(s(\cdot)\). 
% and the score function is defined as \{'Good': 0.9, 'Normal': 0.2, 'Bad': 0.05\}.

The first term encourages the model 
$p_\theta$ to output aligned with the reward signal $r_\phi$ by assigning higher probabilities to high-reward samples from the synthesized dataset 
$\mathcal{D}^{{\mathrm{syn}}}$, where we set the score function \(s(\cdot)\) as \{``Good'': 0.9, ``Normal'': 0.2, ``Bad'': 0.05\}. The second term is designed to mitigate the limitations of training exclusively on the synthesized dataset since synthesized videos often suffer from low temporal consistency, which may hinder the model's ability to maintain subject alignment across frames. 
By incorporating a real video-text dataset $\mathcal{D}^{{\mathrm{real}}}$, the second term acts as a regularizer, grounding the model in realistic frame-to-frame dynamics and ensuring that it learns to generate videos with higher semantic and temporal fidelity. 
The hyperparameter \(\lambda\) balances the loss between synthetic datasets \(\mathcal{D}^{\mathrm{syn}}\) 
and the real dataset \(\mathcal{D}^{\mathrm{real}}\), where we set \(\lambda=1 \) here. 
This balance between synthesized and real data helps the model to generalize better, achieving improved performance in overall video quality.
% This ensures the model can leverage both human-aligned synthetic data and real-world data effectively.
\begin{table*}[!ht]
\setlength\tabcolsep{8.5pt}
\centering
\scriptsize
\caption{\textbf{Quantitative results on video assessment metrics}. The first seven metrics correspond to the \textit{Quality} type, while the remaining correspond to the \textit{Semantic} type. ``RM'' denotes the ``Reward Model''.}

\begin{tabular}{ccccccccccc}

		\toprule
         Models & \makecell[c]{RM \\ Size} &
	\makecell[c]{Subject\\ Consistency}
      & \makecell[c]{Background\\ Consistency}
  &\makecell[c]{Aesthetic \\Quality} &\makecell[c]{Imaging \\Quality}

&\makecell[c]{Temporal \\Flickering}
&\makecell[c]{Motion \\Smoothness}
& \makecell[c]{Dynamic \\Degree}
& \makecell[c]{Human \\Action}

       \\
		\midrule

 		CogVideoX-2B  &     &  94.58       & 95.45     & 61.94  &63.04 & 96.94 &97.86 &60.22 &97.34    \\
        		CogVideoX-5B   &  &   94.84            & 95.64 &63.13    &63.14 &97.19 & 97.81 & \underline{62.78}& \underline{98.40}        \\
		\midrule
                
          \multirow{2}{*}{\makecell[c]{CogVideoX-2B-LiFT \\ \textbf{(Ours)}}} &     13B    & \underline{96.15}             & \underline{96.48}    &\underline{63.24}   &\underline{64.01} &\underline{98.10}& \underline{98.17}& 61.89 & 97.90   \\
           &    40B     & \textbf{96.82}             & \textbf{96.79}    &\textbf{63.72}   &\textbf{64.19} &\textbf{98.20}& \textbf{98.33} & \textbf{62.85} & \textbf{98.44}   \\
		\bottomrule

        \toprule
                
		Models  & \makecell[c]{RM \\ Size} &
		\makecell[c]{Color}
       & \makecell[c]{Spatial \\Relationship}
  &\makecell[c]{Scene} &\makecell[c]{Temporal \\Style}
&\makecell[c]{Overall \\Consistency}
% &\makecell[c]{Temporal \\Flickering}
% &\makecell[c]{Motion \\Smoothness}
% & \makecell[c]{Dynamic \\Degree}
&\makecell[c]{Object \\Class}
&\makecell[c]{Multiple \\Objects}
&\makecell[c]{Appearance \\Style}
       \\
		\midrule

		CogVideoX-2B &     &  80.86     & 61.78     & 53.30  &24.19 & 27.34 &85.75 &69.11 &24.67      \\
        		CogVideoX-5B   &    &   82.50            & 55.94 &56.57    &\underline{25.12} &\underline{27.84} & 88.99 & 70.98& 24.70    \\
		\midrule
                
            \multirow{2}{*}{\makecell[c]{CogVideoX-2B-LiFT \\ \textbf{(Ours)}}}   &  13B    & \underline{84.40}             & \underline{64.66}    &\underline{56.76}   &24.83 &27.23& \underline{91.27} & \underline{77.45} & \underline{24.86} \\
        
               &  40B    & \textbf{85.15}             & \textbf{66.04}    &\textbf{57.63}   &\textbf{25.23} &\textbf{27.93}& \textbf{91.77} & \textbf{79.34} & \textbf{25.92}   \\
		\bottomrule

	\end{tabular} \\
\vspace{-0.1cm}

\label{tab:vbench_compare}
\end{table*}

RWL assigns weights to all samples, allowing the model to leverage the entire dataset. This smooth weighting ensures higher-reward samples contribute more significantly while still utilizing lower-reward samples, enhancing data efficiency and preventing overfitting. The potential problem is the training dataset can become quite large, which may result in increased computational and time costs.
Therefore, we also explore another effective reward-learning function.
\textbf{Rejection Sampling (RS)}.
This method can be viewed as a special case of reward-weighted learning, where the score function serves as a hard filter.
Specifically, the reward weight is defined such that \(r_\phi(\textbf{\textit{x}}, z) = 1\) for samples evaluated as \textit{Good} in all dimensions \textit{D}, and \(r_\phi(\textbf{\textit{x}}, z) = 0\) for all other. 
Under this definition, the loss function for RS simplifies to:
\begin{equation}
\begin{aligned}
\mathcal{L}_{\mathrm{RS}}(\theta) = &\ 
\mathbb{E}_{(\textbf{\textit{x}}, z) \sim \mathcal{D}^{\mathrm{syn}}_{\mathrm{filtered}}}
\left[-\log p_\theta(\textbf{\textit{x}}|z)\right] \\
&+ 
\lambda \cdot \mathbb{E}_{(\textbf{\textit{x}}, z) \sim \mathcal{D}^{\mathrm{real}}}
\left[-\log p_\theta(\textbf{\textit{x}}|z)\right].
\end{aligned}
\end{equation}
Here, \(\mathcal{D}^{\mathrm{syn}}_{\mathrm{filtered}} = \{ (\textbf{\textit{x}}, z) \in \mathcal{D}^{\mathrm{syn}} \,|\, r_{\phi,d}(\textbf{\textit{x}}, z) = Good, d \in D \}\), where $\mathcal{D}^{\mathrm{syn}}$ denotes the original synthesized dataset.

This design effectively reduces the synthesized dataset \(\mathcal{D}^{\mathrm{syn}}\) to only include high-quality samples, achieving a similar effect as applying a binary mask. 

Generally, in comparing RS and RWL, we identify key trade-offs: RS is lightweight and efficient but risks overfitting due to reduced training data, while RWL, though computationally more intensive, leverages all reward samples to enhance data efficiency and prevent overfitting. Our work explores the effectiveness of both methods in experiments.

\begin{figure*}[t]

    \centering
    \includegraphics[width=1\linewidth]{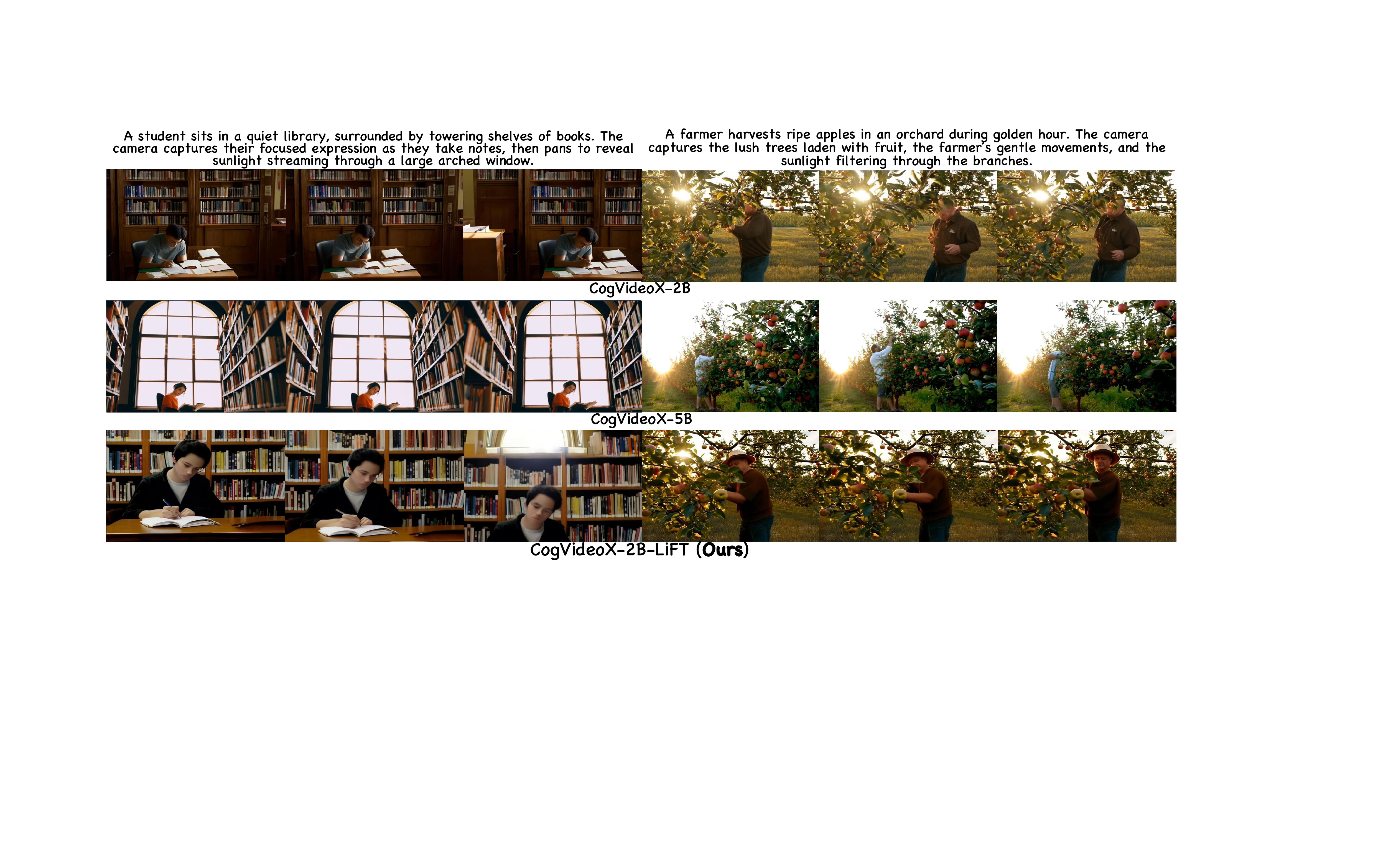}
    \caption{\textbf{Qualitative Comparison}. We compare the performance of our CogVideoX-2B-LiFT (fine-tuned using reward-weighted learning) against CogVideoX-2B and CogVideoX-5B.}
    \label{fig:compare}

\end{figure*}

\section{Experiments}

\subsection{Experimental Setup}
\mypara{Models and Settings.} 
For our reward model \ourmethod, we leverage VILA-1.5 13B/40B \cite{lin2024vila}, which has been pre-trained on extensive video understanding datasets, demonstrating robust capabilities in video comprehension tasks. To adapt the model for our specific evaluation scenario, we employ Low-Rank Adaptation (LoRA) \cite{hu2021lora} to fine-tune all linear layers.
For our baseline T2V generative model, we adopt CogVideoX-2B \cite{yang2024cogvideox}, a foundational T2V generation model. We fine-tune all its transformer blocks to enhance its performance, enabling the model to better align with human preferences and improve video quality. More  implementation details are provided in Appendix \app{B.1}.

% All our models are trained on 8 NVIDIA H100 GPUs with batch size 8 and learning rate 1e$^{-5}$. 

\mypara{Datasets.}
We fine-tune \ourmethod using our proposed \ourdatasetit dataset, which comprises approximately 10K video quality question-answer pairs annotated with human feedback. Of these, 9K samples are used for training, while the remaining samples serve as the test set. Fig. \ref{fig:dataset} visualizes the statistic result of this dataset.
To diversify the dataset and expose the model to real video distributions, we incorporate 1K high-quality real video samples from \textit{VIDGEN} \cite{tan2024vidgen}. For T2V model alignment, we generate 40K videos from prompts using CogVideoX-2B as the synthesized dataset and select approximately 20K video-text pairs from \textit{OpenVid} \cite{nan2024openvid} as the real dataset.

% \mypara{Evaluation Metrics.} we employ Vbench \cite{Huang2023VBenchCB}, a comprehensive benchmark suite to assess the performance of T2V generation, which decomposes ``video generation quality'' into specific, hierarchical, and disentangled dimensions. Each dimension is evaluated using tailored prompts and specialized evaluation methods. 
% The evaluation prompts are optimized using Qwen2.5-72B-Instruct \cite{yang2024qwen2} since the CogVideoX \cite{yang2024cogvideox} model is trained with long prompts.

% \input{tables/dataset_distribution}
% \subsection{Reward Model Baseline} VideoScore, VBench, EvalCrafter, GenAI-Bench

\begin{table}[t]
\setlength
\tabcolsep{5.5pt}
\centering
\scriptsize
\caption{\textbf{Evaluation results of our reward model}. We assess \ourmethod across the first three dimensions in the test set and compute the average accuracy based on these evaluations.}

\begin{tabular}{ccccccc}

		\toprule
          & Size &
	\makecell[c]{Semantic\\ Consistency}
      & \makecell[c]{Motion\\ Smoothness}
  &\makecell[c]{Video \\Fidelity} &\makecell[c]{Avg. \\Accuracy}

       \\

       		\midrule
            \rowcolor{gray!20}
    GPT-4o-mini  &     &  60.42       & 64.33     &  68.26 &  64.34   \\
  
    \rowcolor{gray!20}
    GPT-4o  &   &    70.59       &   67.29   &  71.99 & 69.96\\
		\midrule
 		LiFT-Critic  & 13B    &  80.27       & 77.48     & 82.72  &80.15     \\

		 w/o reason & 40B &   84.55            & 88.01 &88.62    &87.06         \\

		\midrule

 		\multirow{2}{*}{LiFT-Critic}  & 13B    &  85.61       & 88.92     & 88.51  &87.68     \\

		  & 40B &   \textbf{90.23}            & \textbf{94.89} &\textbf{95.53}    &\textbf{93.55}         \\
		\bottomrule

	\end{tabular} \\

    \vspace{-0.2cm}
\label{tab:reward_size}
\end{table}

\begin{figure*}[!thb]

    \centering
    \includegraphics[width=0.9\linewidth]{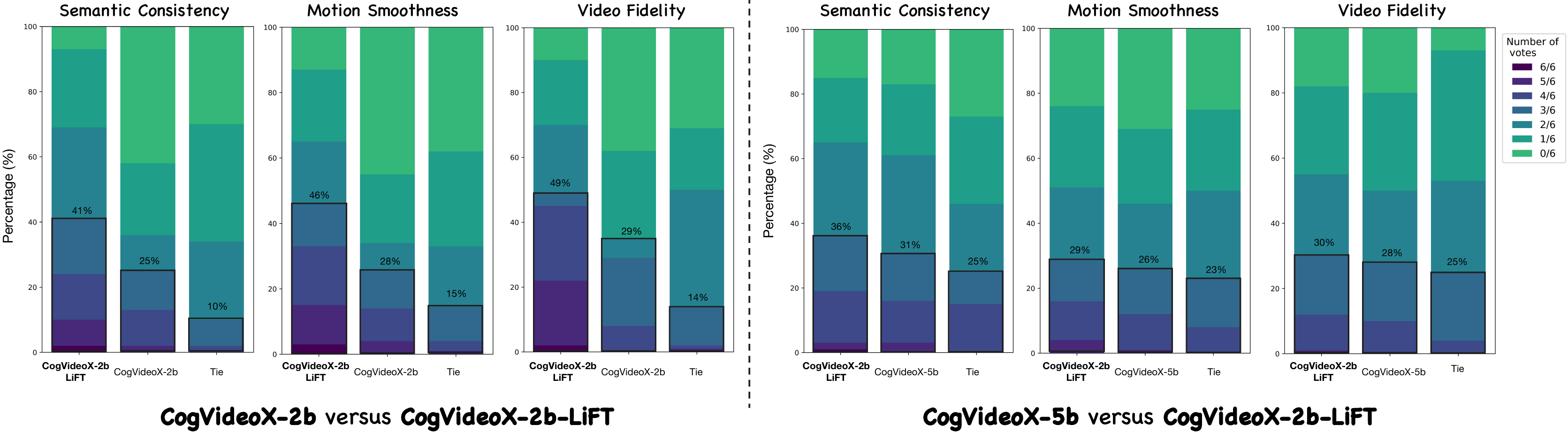}
    \caption{\textbf{Human evaluation results}. We report the percentage of queries receiving positive votes for each method and highlight the percentage of queries achieving a majority consensus (at least three out of six positive votes) in the black box.}
    \label{fig:human_evaluation}
    \vspace{-0.3cm}
\end{figure*}

\begin{figure*}[t]

    \centering
    \includegraphics[width=0.95\linewidth]{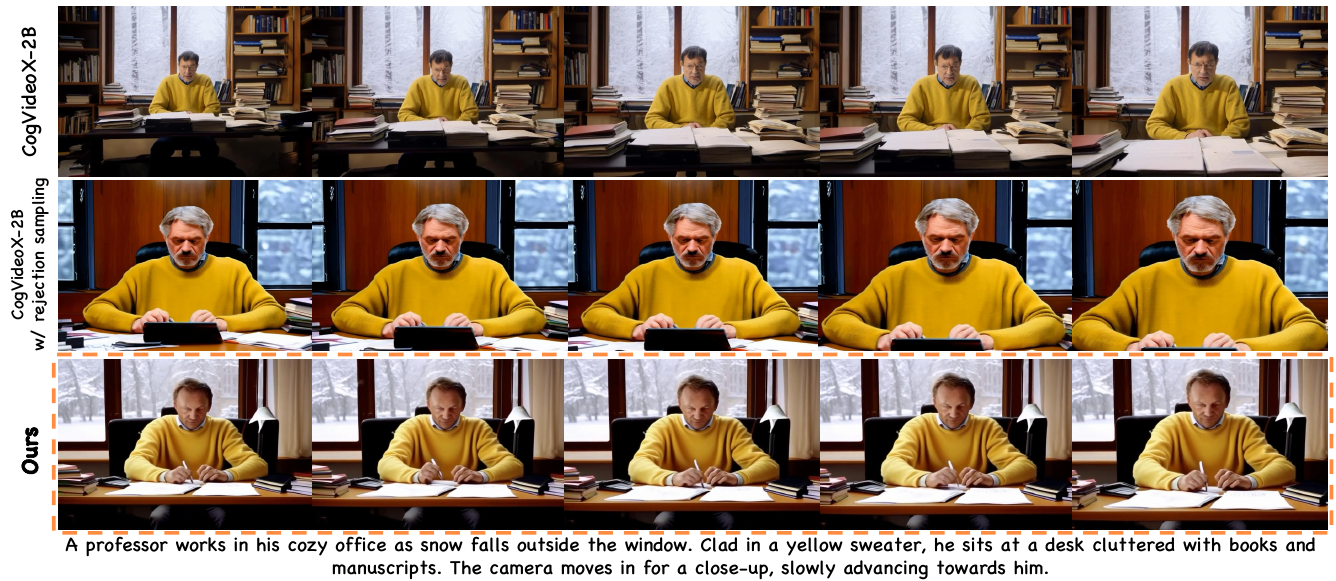}
    \caption{\textbf{Qualitative comparison}. We compare the performance of CogVideo-2B, its variations fine-tuned using reward-weighted learning (Ours) and rejection sampling.}
    \label{fig:compare_rejection}
    % \vspace{-0.1cm}
\end{figure*}

\begin{figure}[t]

    \centering
    \includegraphics[width=0.85\linewidth]{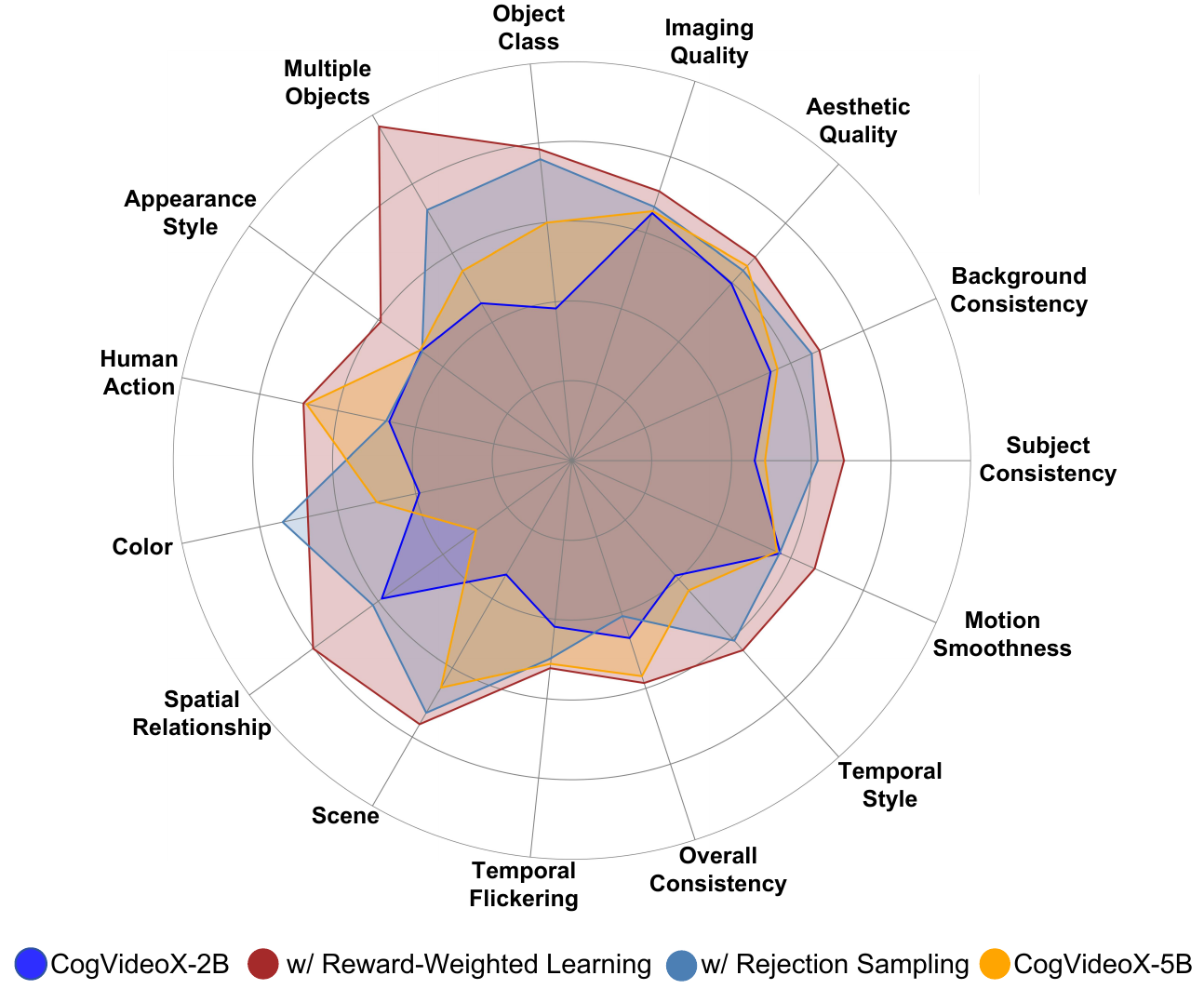}
    \caption{\textbf{Visualized evaluation results in multiple evaluation dimensions}. The middle two methods in the label region represent the CogVideoX-2B model fine-tuned using different reward learning strategies.}
    \label{fig:radar}
    \vspace{-0.1cm}

\end{figure}

\subsection{Results}
\mypara{Quantitative Results.} Tab. \ref{tab:vbench_compare} demonstrates the significant performance improvements achieved by our fine-tuning pipeline. Specifically, integrating reward learning using \ourmethod at various scales leads to consistent enhancements across nearly all evaluation metrics. For example, the model fine-tuned with \ourmethod-40B shows marked improvements in ``Subject Consistency'' and ``Motion Smoothness'', reflecting better alignment with human preferences for coherent video generation and fluid motion. This enhancement also indicates that \ourmethod-40B better understands and captures the continuity and smoothness required in high-quality video.
When compared to the larger baseline, CogVideoX-5B, the \ourmethod-40B-enhanced model outperforms in critical areas such as ``Imaging Quality'' and ``Multiple Objects''. This improvement highlights the model’s superior ability to generate visually detailed scenes with multiple objects, which is essential for creating richer and more complex video content. We also apply our pipeline to T2V-turbo \cite{Li2024T2VTurboBT}, and the results are shown in Appendix \app{B.4}. 
% Furthermore, \ourmethod-40B remains competitive in aspects such as "Temporal Style," showcasing its continued ability to maintain consistent dynamic properties over time.

% These results emphasize the impact of human feedback-driven reward learning, which not only refines subjective qualities like smoothness and style but also strengthens objective attributes such as spatial consistency and object representation. 
\mypara{Qualitative Results.} Fig. \ref{fig:compare} displays the visual comparison results. It is clear that our fine-tuned model achieves relatively better performance in terms of semantic consistency, motion smoothness, and video fidelity. Specifically, in the first case, CogVideoX-2B fails to perform the required camera transition as described in the caption, while our method successfully captures this motion. In the second case, both CogVideoX-2B and CogVideoX-5B generate a farmer with a very blurred face, whereas our model produces a clearer and more detailed facial representation. These examples highlight the effectiveness of our approach in aligning the generated video with both the textual description and human expectations. More qualitative comparison results are provided in Appendix \app{B.5}.

The quantitative comparison and qualitative results of our \ourmethod are provided in Appendix. \app{B.2-3}.

% Overall, both quantitative and qualitative results emphasize that our \ourmethod is highly effective in aligning T2V models with human expectations, achieving comprehensive improvements across a broad range of evaluation criteria and visual quality.

\subsection{Human Evaluation}
We conduct a human evaluation to compare CogVideoX-2B, CogVideoX-5B, and our CogVideoX-2B-LiFT. Using a set of 120 text prompts, we generate videos for each method to ensure consistency.
To assess performance, we perform pairwise comparisons between CogVideoX-2B-LiFT and both CogVideoX-2B and CogVideoX-5B. Human raters evaluate each pair, selecting the better-performing method or indicating a tie if both produce similar results. The results, summarized in Fig. \ref{fig:human_evaluation}, show the percentage of queries receiving positive votes for each method. We also highlight the proportion of cases achieving a majority consensus (at least three out of six positive votes) in the black box. These results demonstrate the advantages of CogVideoX-2B-LiFT over its counterparts. See Appendix. \app{B.6} for more details.
\subsection{Ablation Studies} 

\mypara{Choices of reward learning.} \label{sec:choice}
As introduced in Sec. \ref{sec:align}, we also explore another effective reward learning function, rejection sampling (RS). In this process, only synthesized videos that receive a ``Good'' score across all three evaluation dimensions are retained. Compared to reward-weighted learning (RWL), this approach effectively reduces the amount of fine-tuning data while ensuring the selected samples meet high-quality standards for model alignment.
Fig. \ref{fig:radar} presents a comprehensive overview of performance across multiple evaluation dimensions. The results show that RS significantly outperforms the baseline CogVideoX-2B, while maintaining competitive performance with the larger CogVideoX-5B. Although slightly less effective than RWL, RS proves to be a more efficient and lightweight method for aligning T2V models with human preferences. Additionally, Fig. \ref{fig:compare_rejection} showcases the visual improvements and enhanced alignment with human expectations achieved through these approaches.
% \subsubsection{Effects of reward learning dataset size}

\mypara{Effects of reward model size}.
We explore the performance comparison of \ourmethod at various scales. Tab. \ref{tab:reward_size} demonstrates that increasing the model size significantly improves performance across all evaluation metrics, resulting in a notable increase in overall accuracy.  Besides, the results in Tab. \ref{tab:vbench_compare} illustrate the impact of reward model size on T2V model performance. Compared to the 13B, the \ourmethod-40B enhanced model achieves consistent improvements across all metrics. These enhancements indicate that a larger reward model better captures nuanced human preferences, leading to improved alignment with subjective quality assessments.

\mypara{Effects of learning reason}. 
Tab. \ref{tab:reward_size} underscores the advantages of incorporating reason-based annotations into reward function learning. By comparing \ourmethod with and without reasoning at the same model size, we observe significant performance improvements. Notably, for the 40B model, the inclusion of reason boosts the average accuracy from 87.06\% to 93.55\%. These results emphasize that providing reasoning behind human feedback allows the reward model to better capture subtle evaluation criteria, leading to improved alignment with human preferences.

The effect of the real video-text dataset in reward learning is analyzed in Appendix. \app{B.7}.

\section{Conclusion}
This work proposes the first fine-tuning pipeline \ourpipeline for aligning T2V models with human preferences. First, we curate a human-annotated dataset, \ourdataset, which includes both ratings and reason for video evaluation. We then train a reward model \ourmethod to learn the reward function from this dataset. Lastly, we employ it to align the T2V model. As a case study, we apply this pipeline to CogVideoX-2B, showing that the fine-tuned model outperforms the CogVideoX-5B across all 16 metrics, showcasing the effectiveness of our approach in aligning T2V models with human expectations. 
% We believe our method provides valuable insights and will contribute to advancing the development of more human preference-aligned T2V generation systems.

{
    \small
    \bibliographystyle{ieeenat_fullname}
    \bibliography{egbib}

\begin{thebibliography}{42}
\providecommand{\natexlab}[1]{#1}
\providecommand{\url}[1]{\texttt{#1}}
\expandafter\ifx\csname urlstyle\endcsname\relax
  \providecommand{\doi}[1]{doi: #1}\else
  \providecommand{\doi}{doi: \begingroup \urlstyle{rm}\Url}\fi

\bibitem[Bai et~al.(2022)Bai, Jones, Ndousse, Askell, Chen, DasSarma, Drain, Fort, Ganguli, Henighan, et~al.]{bai2022training}
Yuntao Bai, Andy Jones, Kamal Ndousse, Amanda Askell, Anna Chen, Nova DasSarma, Dawn Drain, Stanislav Fort, Deep Ganguli, Tom Henighan, et~al.
\newblock Training a helpful and harmless assistant with reinforcement learning from human feedback.
\newblock \emph{arXiv preprint arXiv:2204.05862}, 2022.

\bibitem[Black et~al.(2023)Black, Janner, Du, Kostrikov, and Levine]{black2023training}
Kevin Black, Michael Janner, Yilun Du, Ilya Kostrikov, and Sergey Levine.
\newblock Training diffusion models with reinforcement learning.
\newblock \emph{arXiv preprint arXiv:2305.13301}, 2023.

\bibitem[Blattmann et~al.(2023)Blattmann, Rombach, Ling, Dockhorn, Kim, Fidler, and Kreis]{blattmann2023align}
Andreas Blattmann, Robin Rombach, Huan Ling, Tim Dockhorn, Seung~Wook Kim, Sanja Fidler, and Karsten Kreis.
\newblock Align your latents: High-resolution video synthesis with latent diffusion models.
\newblock In \emph{CVPR}, pages 22563--22575, 2023.

\bibitem[Chen et~al.(2025)Chen, Wang, Wu, Liao, Sun, Yan, and Lin]{chen2025enhancing}
Chaofeng Chen, Annan Wang, Haoning Wu, Liang Liao, Wenxiu Sun, Qiong Yan, and Weisi Lin.
\newblock Enhancing diffusion models with text-encoder reinforcement learning.
\newblock In \emph{ECCV}, pages 182--198, 2025.

\bibitem[Chen et~al.(2023{\natexlab{a}})Chen, Xia, He, Zhang, Cun, Yang, Xing, Liu, Chen, Wang, et~al.]{chen2023videocrafter1}
Haoxin Chen, Menghan Xia, Yingqing He, Yong Zhang, Xiaodong Cun, Shaoshu Yang, Jinbo Xing, Yaofang Liu, Qifeng Chen, Xintao Wang, et~al.
\newblock Videocrafter1: Open diffusion models for high-quality video generation.
\newblock \emph{arXiv preprint arXiv:2310.19512}, 2023{\natexlab{a}}.

\bibitem[Chen et~al.(2023{\natexlab{b}})Chen, Xu, Ren, Cong, He, Xie, Sinha, Luo, Xiang, and Perez-Rua]{chen2023gentron}
Shoufa Chen, Mengmeng Xu, Jiawei Ren, Yuren Cong, Sen He, Yanping Xie, Animesh Sinha, Ping Luo, Tao Xiang, and Juan-Manuel Perez-Rua.
\newblock Gentron: Delving deep into diffusion transformers for image and video generation.
\newblock \emph{arXiv preprint arXiv:2312.04557}, 2023{\natexlab{b}}.

\bibitem[Chen et~al.(2024)Chen, Wu, Wang, Su, Chen, Xing, Zhong, Zhang, Zhu, Lu, et~al.]{chen2024internvl}
Zhe Chen, Jiannan Wu, Wenhai Wang, Weijie Su, Guo Chen, Sen Xing, Muyan Zhong, Qinglong Zhang, Xizhou Zhu, Lewei Lu, et~al.
\newblock Internvl: Scaling up vision foundation models and aligning for generic visual-linguistic tasks.
\newblock In \emph{CVPR}, pages 24185--24198, 2024.

\bibitem[Clark et~al.(2023)Clark, Vicol, Swersky, and Fleet]{Clark2023DirectlyFD}
Kevin Clark, Paul Vicol, Kevin Swersky, and David~J. Fleet.
\newblock Directly fine-tuning diffusion models on differentiable rewards.
\newblock \emph{arXiv preprint arXiv:2309.17400}, 2023.

\bibitem[Fan et~al.(2024)Fan, Watkins, Du, Liu, Ryu, Boutilier, Abbeel, Ghavamzadeh, Lee, and Lee]{fan2024reinforcement}
Ying Fan, Olivia Watkins, Yuqing Du, Hao Liu, Moonkyung Ryu, Craig Boutilier, Pieter Abbeel, Mohammad Ghavamzadeh, Kangwook Lee, and Kimin Lee.
\newblock Reinforcement learning for fine-tuning text-to-image diffusion models.
\newblock \emph{NeurIPS}, 36, 2024.

\bibitem[Guo et~al.(2023)Guo, Yang, Rao, Liang, Wang, Qiao, Agrawala, Lin, and Dai]{guo2023animatediff}
Yuwei Guo, Ceyuan Yang, Anyi Rao, Zhengyang Liang, Yaohui Wang, Yu Qiao, Maneesh Agrawala, Dahua Lin, and Bo Dai.
\newblock Animatediff: Animate your personalized text-to-image diffusion models without specific tuning.
\newblock \emph{arXiv preprint arXiv:2307.04725}, 2023.

\bibitem[Gupta et~al.(2025)Gupta, Yu, Sohn, Gu, Hahn, Li, Essa, Jiang, and Lezama]{gupta2025photorealistic}
Agrim Gupta, Lijun Yu, Kihyuk Sohn, Xiuye Gu, Meera Hahn, Fei-Fei Li, Irfan Essa, Lu Jiang, and Jos{\'e} Lezama.
\newblock Photorealistic video generation with diffusion models.
\newblock In \emph{ECCV}, pages 393--411, 2025.

\bibitem[He et~al.(2024{\natexlab{a}})He, Xue, Liu, Lin, Gao, Lin, Qiao, Ouyang, and Liu]{he2024venhancer}
Jingwen He, Tianfan Xue, Dongyang Liu, Xinqi Lin, Peng Gao, Dahua Lin, Yu Qiao, Wanli Ouyang, and Ziwei Liu.
\newblock Venhancer: Generative space-time enhancement for video generation.
\newblock \emph{arXiv preprint arXiv:2407.07667}, 2024{\natexlab{a}}.

\bibitem[He et~al.(2024{\natexlab{b}})He, Jiang, Zhang, Ku, Soni, Siu, Chen, Chandra, Jiang, Arulraj, et~al.]{he2024videoscore}
Xuan He, Dongfu Jiang, Ge Zhang, Max Ku, Achint Soni, Sherman Siu, Haonan Chen, Abhranil Chandra, Ziyan Jiang, Aaran Arulraj, et~al.
\newblock Videoscore: Building automatic metrics to simulate fine-grained human feedback for video generation.
\newblock \emph{arXiv preprint arXiv:2406.15252}, 2024{\natexlab{b}}.

\bibitem[Hu et~al.(2021)Hu, Shen, Wallis, Allen-Zhu, Li, Wang, Wang, and Chen]{hu2021lora}
Edward~J Hu, Yelong Shen, Phillip Wallis, Zeyuan Allen-Zhu, Yuanzhi Li, Shean Wang, Lu Wang, and Weizhu Chen.
\newblock Lora: Low-rank adaptation of large language models.
\newblock \emph{arXiv preprint arXiv:2106.09685}, 2021.

\bibitem[Huang et~al.(2023)Huang, He, Yu, Zhang, Si, Jiang, Zhang, Wu, Jin, Chanpaisit, Wang, Chen, Wang, Lin, Qiao, and Liu]{Huang2023VBenchCB}
Ziqi Huang, Yinan He, Jiashuo Yu, Fan Zhang, Chenyang Si, Yuming Jiang, Yuanhan Zhang, Tianxing Wu, Qingyang Jin, Nattapol Chanpaisit, Yaohui Wang, Xinyuan Chen, Limin Wang, Dahua Lin, Yu Qiao, and Ziwei Liu.
\newblock Vbench: Comprehensive benchmark suite for video generative models.
\newblock \emph{CVPR}, pages 21807--21818, 2023.

\bibitem[Kirstain et~al.(2023)Kirstain, Polyak, Singer, Matiana, Penna, and Levy]{kirstain2023pick}
Yuval Kirstain, Adam Polyak, Uriel Singer, Shahbuland Matiana, Joe Penna, and Omer Levy.
\newblock Pick-a-pic: An open dataset of user preferences for text-to-image generation.
\newblock \emph{NeurIPS}, 36:\penalty0 36652--36663, 2023.

\bibitem[Kou et~al.(2024)Kou, Liu, Zhang, Li, Wu, Min, Zhai, and Liu]{Kou2024SubjectiveAlignedDA}
Tengchuan Kou, Xiaohong Liu, Zicheng Zhang, Chunyi Li, Haoning Wu, Xiongkuo Min, Guangtao Zhai, and Ning Liu.
\newblock Subjective-aligned dataset and metric for text-to-video quality assessment.
\newblock \emph{arXiv preprint arXiv:2403.11956}, 2024.

\bibitem[Lee et~al.(2023)Lee, Liu, Ryu, Watkins, Du, Boutilier, Abbeel, Ghavamzadeh, and Gu]{lee2023aligning}
Kimin Lee, Hao Liu, Moonkyung Ryu, Olivia Watkins, Yuqing Du, Craig Boutilier, Pieter Abbeel, Mohammad Ghavamzadeh, and Shixiang~Shane Gu.
\newblock Aligning text-to-image models using human feedback.
\newblock \emph{arXiv preprint arXiv:2302.12192}, 2023.

\bibitem[Li et~al.(2023)Li, Zhang, Wu, Sun, Min, Liu, Zhai, and Lin]{li2023agiqa}
Chunyi Li, Zicheng Zhang, Haoning Wu, Wei Sun, Xiongkuo Min, Xiaohong Liu, Guangtao Zhai, and Weisi Lin.
\newblock Agiqa-3k: An open database for ai-generated image quality assessment.
\newblock \emph{IEEE Transactions on Circuits and Systems for Video Technology}, 2023.

\bibitem[Li et~al.(2024{\natexlab{a}})Li, Feng, Chen, and Wang]{Li2024RewardGL}
Jiachen Li, Weixi Feng, Wenhu Chen, and William~Yang Wang.
\newblock Reward guided latent consistency distillation.
\newblock \emph{arXiv preprint arXiv:2403.11027}, 2024{\natexlab{a}}.

\bibitem[Li et~al.(2024{\natexlab{b}})Li, Feng, Fu, Wang, Basu, Chen, and Wang]{Li2024T2VTurboBT}
Jiachen Li, Weixi Feng, Tsu-Jui Fu, Xinyi Wang, Sugato Basu, Wenhu Chen, and William~Yang Wang.
\newblock T2v-turbo: Breaking the quality bottleneck of video consistency model with mixed reward feedback.
\newblock \emph{arXiv preprint arXiv:2405.18750}, 2024{\natexlab{b}}.

\bibitem[Li et~al.(2024{\natexlab{c}})Li, Long, Zheng, Gao, Piramuthu, Chen, and Wang]{li2024t2v}
Jiachen Li, Qian Long, Jian Zheng, Xiaofeng Gao, Robinson Piramuthu, Wenhu Chen, and William~Yang Wang.
\newblock T2v-turbo-v2: Enhancing video generation model post-training through data, reward, and conditional guidance design.
\newblock \emph{arXiv preprint arXiv:2410.05677}, 2024{\natexlab{c}}.

\bibitem[Liang et~al.(2024)Liang, He, Li, Li, Klimovskiy, Carolan, Sun, Pont-Tuset, Young, Yang, et~al.]{liang2024rich}
Youwei Liang, Junfeng He, Gang Li, Peizhao Li, Arseniy Klimovskiy, Nicholas Carolan, Jiao Sun, Jordi Pont-Tuset, Sarah Young, Feng Yang, et~al.
\newblock Rich human feedback for text-to-image generation.
\newblock In \emph{CVPR}, pages 19401--19411, 2024.

\bibitem[Lin et~al.(2024)Lin, Yin, Ping, Molchanov, Shoeybi, and Han]{lin2024vila}
Ji Lin, Hongxu Yin, Wei Ping, Pavlo Molchanov, Mohammad Shoeybi, and Song Han.
\newblock Vila: On pre-training for visual language models.
\newblock In \emph{CVPR}, pages 26689--26699, 2024.

\bibitem[Nan et~al.(2024)Nan, Xie, Zhou, Fan, Yang, Chen, Li, Yang, and Tai]{nan2024openvid}
Kepan Nan, Rui Xie, Penghao Zhou, Tiehan Fan, Zhenheng Yang, Zhijie Chen, Xiang Li, Jian Yang, and Ying Tai.
\newblock Openvid-1m: A large-scale high-quality dataset for text-to-video generation.
\newblock \emph{arXiv preprint arXiv:2407.02371}, 2024.

\bibitem[Ouyang et~al.(2022)Ouyang, Wu, Jiang, Almeida, Wainwright, Mishkin, Zhang, Agarwal, Slama, Ray, et~al.]{ouyang2022training}
Long Ouyang, Jeffrey Wu, Xu Jiang, Diogo Almeida, Carroll Wainwright, Pamela Mishkin, Chong Zhang, Sandhini Agarwal, Katarina Slama, Alex Ray, et~al.
\newblock Training language models to follow instructions with human feedback.
\newblock \emph{NeurIPS}, 35:\penalty0 27730--27744, 2022.

\bibitem[Prabhudesai et~al.(2024)Prabhudesai, Mendonca, Qin, Fragkiadaki, and Pathak]{prabhudesai2024video}
Mihir Prabhudesai, Russell Mendonca, Zheyang Qin, Katerina Fragkiadaki, and Deepak Pathak.
\newblock Video diffusion alignment via reward gradients.
\newblock \emph{arXiv preprint arXiv:2407.08737}, 2024.

\bibitem[Radford et~al.(2021)Radford, Kim, Hallacy, Ramesh, Goh, Agarwal, Sastry, Askell, Mishkin, Clark, et~al.]{radford2021learning}
Alec Radford, Jong~Wook Kim, Chris Hallacy, Aditya Ramesh, Gabriel Goh, Sandhini Agarwal, Girish Sastry, Amanda Askell, Pamela Mishkin, Jack Clark, et~al.
\newblock Learning transferable visual models from natural language supervision.
\newblock In \emph{ICML}, pages 8748--8763, 2021.

\bibitem[Tan et~al.(2024)Tan, Yang, Qin, and Li]{tan2024vidgen}
Zhiyu Tan, Xiaomeng Yang, Luozheng Qin, and Hao Li.
\newblock Vidgen-1m: A large-scale dataset for text-to-video generation.
\newblock \emph{arXiv preprint arXiv:2408.02629}, 2024.

\bibitem[Wang et~al.(2023{\natexlab{a}})Wang, Yuan, Chen, Zhang, Wang, and Zhang]{wang2023modelscope}
Jiuniu Wang, Hangjie Yuan, Dayou Chen, Yingya Zhang, Xiang Wang, and Shiwei Zhang.
\newblock Modelscope text-to-video technical report.
\newblock \emph{arXiv preprint arXiv:2308.06571}, 2023{\natexlab{a}}.

\bibitem[Wang et~al.(2024)Wang, Yuan, Zhang, Chen, Wang, Zhang, Shen, Zhao, and Zhou]{wang2024videocomposer}
Xiang Wang, Hangjie Yuan, Shiwei Zhang, Dayou Chen, Jiuniu Wang, Yingya Zhang, Yujun Shen, Deli Zhao, and Jingren Zhou.
\newblock Videocomposer: Compositional video synthesis with motion controllability.
\newblock \emph{NeurIPS}, 36, 2024.

\bibitem[Wang et~al.(2023{\natexlab{b}})Wang, Chen, Ma, Zhou, Huang, Wang, Yang, He, Yu, Yang, et~al.]{wang2023lavie}
Yaohui Wang, Xinyuan Chen, Xin Ma, Shangchen Zhou, Ziqi Huang, Yi Wang, Ceyuan Yang, Yinan He, Jiashuo Yu, Peiqing Yang, et~al.
\newblock Lavie: High-quality video generation with cascaded latent diffusion models.
\newblock \emph{arXiv preprint arXiv:2309.15103}, 2023{\natexlab{b}}.

\bibitem[Wu et~al.(2023)Wu, Hao, Sun, Chen, Zhu, Zhao, and Li]{wu2023human}
Xiaoshi Wu, Yiming Hao, Keqiang Sun, Yixiong Chen, Feng Zhu, Rui Zhao, and Hongsheng Li.
\newblock Human preference score v2: A solid benchmark for evaluating human preferences of text-to-image synthesis.
\newblock \emph{arXiv preprint arXiv:2306.09341}, 2023.

\bibitem[Xu et~al.(2024)Xu, Liu, Wu, Tong, Li, Ding, Tang, and Dong]{xu2024imagereward}
Jiazheng Xu, Xiao Liu, Yuchen Wu, Yuxuan Tong, Qinkai Li, Ming Ding, Jie Tang, and Yuxiao Dong.
\newblock Imagereward: Learning and evaluating human preferences for text-to-image generation.
\newblock \emph{NeurIPS}, 36, 2024.

\bibitem[Yang et~al.(2024{\natexlab{a}})Yang, Yang, Hui, Zheng, Yu, Zhou, Li, Li, Liu, Huang, et~al.]{yang2024qwen2}
An Yang, Baosong Yang, Binyuan Hui, Bo Zheng, Bowen Yu, Chang Zhou, Chengpeng Li, Chengyuan Li, Dayiheng Liu, Fei Huang, et~al.
\newblock Qwen2 technical report.
\newblock \emph{arXiv preprint arXiv:2407.10671}, 2024{\natexlab{a}}.

\bibitem[Yang et~al.(2024{\natexlab{b}})Yang, Teng, Zheng, Ding, Huang, Xu, Yang, Hong, Zhang, Feng, et~al.]{yang2024cogvideox}
Zhuoyi Yang, Jiayan Teng, Wendi Zheng, Ming Ding, Shiyu Huang, Jiazheng Xu, Yuanming Yang, Wenyi Hong, Xiaohan Zhang, Guanyu Feng, et~al.
\newblock Cogvideox: Text-to-video diffusion models with an expert transformer.
\newblock \emph{arXiv preprint arXiv:2408.06072}, 2024{\natexlab{b}}.

\bibitem[Yuan et~al.(2023)Yuan, Zhang, Wang, Wei, Feng, Pan, Zhang, Liu, Albanie, and Ni]{Yuan2023InstructVideoIV}
Hangjie Yuan, Shiwei Zhang, Xiang Wang, Yujie Wei, Tao Feng, Yining Pan, Yingya Zhang, Ziwei Liu, Samuel Albanie, and Dong Ni.
\newblock Instructvideo: Instructing video diffusion models with human feedback.
\newblock \emph{CVPR}, pages 6463--6474, 2023.

\bibitem[Yuan et~al.(2024)Yuan, Zhang, Wang, Wei, Feng, Pan, Zhang, Liu, Albanie, and Ni]{yuan2024instructvideo}
Hangjie Yuan, Shiwei Zhang, Xiang Wang, Yujie Wei, Tao Feng, Yining Pan, Yingya Zhang, Ziwei Liu, Samuel Albanie, and Dong Ni.
\newblock Instructvideo: instructing video diffusion models with human feedback.
\newblock In \emph{CVPR}, pages 6463--6474, 2024.

\bibitem[Zhai et~al.(2023)Zhai, Mustafa, Kolesnikov, and Beyer]{zhai2023sigmoid}
Xiaohua Zhai, Basil Mustafa, Alexander Kolesnikov, and Lucas Beyer.
\newblock Sigmoid loss for language image pre-training.
\newblock In \emph{ICCV}, pages 11975--11986, 2023.

\bibitem[Zhang et~al.(2024{\natexlab{a}})Zhang, Wang, Wu, Li, Gao, Zhang, and Wang]{zhang2024learning}
Sixian Zhang, Bohan Wang, Junqiang Wu, Yan Li, Tingting Gao, Di Zhang, and Zhongyuan Wang.
\newblock Learning multi-dimensional human preference for text-to-image generation.
\newblock In \emph{CVPR}, pages 8018--8027, 2024{\natexlab{a}}.

\bibitem[Zhang et~al.(2024{\natexlab{b}})Zhang, Tzeng, Du, and Kislyuk]{Zhang2024LargescaleRL}
Yinan Zhang, Eric Tzeng, Yilun Du, and Dmitry Kislyuk.
\newblock Large-scale reinforcement learning for diffusion models.
\newblock \emph{arXiv preprint arXiv:2401.12244}, 2024{\natexlab{b}}.

\bibitem[Zhou et~al.(2024)Zhou, Wang, Cai, and Yang]{zhou2024allegro}
Yuan Zhou, Qiuyue Wang, Yuxuan Cai, and Huan Yang.
\newblock Allegro: Open the black box of commercial-level video generation model.
\newblock \emph{arXiv preprint arXiv:2410.15458}, 2024.

\end{thebibliography}
}

\clearpage 
\appendix
% 添加个list

\begin{table*}[th!]
\centering
\setlength{\tabcolsep}{6pt} % 调整列间距
\renewcommand{\arraystretch}{1.2} % 调整行高
\caption{\textbf{Evaluation Dimensions and Definitions}. This table provides the detailed definitions of each evaluation dimension used in our annotation process.}
\label{tab:evaluation_dimensions}
\begin{tabular}{p{4cm}p{12cm}}
\toprule
\textbf{Evaluation Dimension} & \textbf{Definition} \\ \hline
Semantic Consistency & The consistency and coherence of semantic elements across frames, ensuring the depicted objects, characters, and actions remain contextually appropriate and logically related throughout the video. \\ \hline
Motion Smoothness & The naturalness and fluidity of motion transitions between consecutive frames, reflecting the absence of jittering, abrupt movements, or frame discontinuities in the video. \\ \hline
Video Fidelity & The overall visual accuracy and realism of the video, including the sharpness, texture details, and absence of visual artifacts such as blurriness, distortions, or unnatural effects. \\ \bottomrule

\end{tabular}
\end{table*}

\section{\ourdataset}
\subsection{Video-Text Dataset Generation}
To create a comprehensive video-text dataset, we begin by generating a diverse set of prompts. This process starts with creating selection lists for different categories, including humans, animals, places, and actions. A full list of our categories, along with examples, is provided in Tab. \ref{tab:category}. For each prompt, we randomly choose 1-2 subjects from the human and animal categories, a scene from the places category, and an action from either the simple or complex actions categories. These selected elements are then combined into a phrase, which is further refined into a detailed textual description using a large language model (LLM) \cite{yang2024qwen2}.
Once the prompts are generated, multiple videos are generated for each prompt using T2V models, resulting in a video-text dataset that captures a wide variety of subjects, scenes, and actions.

\subsection{Annotator Management}
We enlist 10 expert annotators, all of whom are undergraduate or graduate students. Some of them will become co-authors of the paper, while others are compensated with a fair salary. All the annotators are proficient in English, ensuring they can fully understand the text prompts.

\subsection{Annotation Guideline}
We categorize human preferences in videos into three key dimensions: semantic consistency, motion smoothness, and video fidelity. The definitions of these dimensions are detailed in Tab. \ref{tab:evaluation_dimensions}. 
For each video-text pair, annotators are asked to evaluate the synthesized videos across these dimensions by assigning a score and providing the corresponding reason. We provide the detailed scoring criteria for each dimension:

\mypara{Semantic Consistency}
\begin{itemize}
    \item \textbf{Good}: Perfectly aligned. The video content matches the given caption with no ambiguity or confusion.
    \item \textbf{Normal}: Minor inconsistencies, but the overall meaning is still clear and understandable.
    \item \textbf{Bad}: Largely inconsistent, with significant discrepancies between the video and the caption, making it hard to interpret the intended meaning.
\end{itemize}

\begin{figure}[h]

    \centering
    \includegraphics[width=0.9\linewidth]{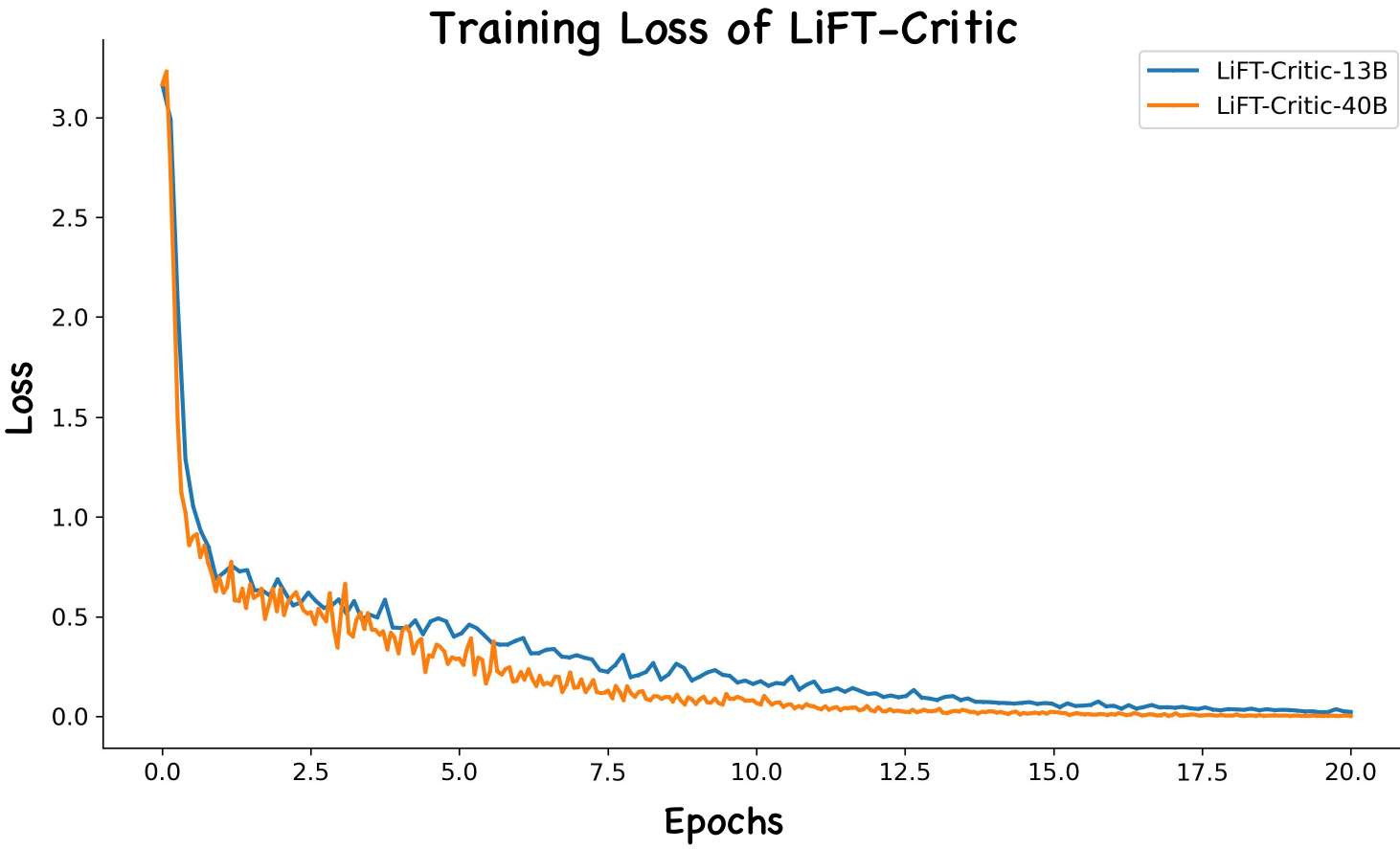}
    \caption{\textbf{Visualization of training loss}. This figure depicts the progression of training loss for \ourmethod 13B/40B over the course of epochs.}
    \label{fig:loss_plot}

\end{figure}

\mypara{Motion Smoothness}
\begin{itemize}
    \item \textbf{Good}: Motions are fluid and natural, with no abrupt transitions or noticeable stuttering.
    \item \textbf{Normal}: Slight motion irregularities or mild jittering that does not distract from the overall visual flow.
    \item \textbf{Bad}: Motions are choppy or erratic, with obvious frame drops or unnatural movements that break the immersion.
\end{itemize}

\mypara{Video Fidelity}
\begin{itemize}
    \item \textbf{Good}: High visual quality, with clear details, sharp images, and minimal artifacts. The video appears natural and realistic.
    \item \textbf{Normal}: Some visual artifacts, such as blurring or slight color inaccuracies, but they do not significantly detract from the video’s quality.
    \item \textbf{Bad}: Low visual quality, with severe blurring, significant color distortion, or noticeable artifacts that impact the viewing experience.
\end{itemize}

To ensure the annotators clearly understand the definitions, we provide a set of several example cases for each rating category in the guideline. Some prompts may contain unfamiliar terms or names, which could lead to confusion. In such cases, we instruct annotators to quickly search for any unknown concepts online and skip samples that contain confusing or obscure terminology.

\subsection{Annotation Interface}
We design a web-based user interface to streamline the data collection process, as shown in Fig. 3 in the main paper. The main interface features the video on the left side, with a panel on the right displaying the corresponding text prompt at the top. For each dimension of evaluation, annotators are first asked to select a score from the available options and then provide a detailed reason in the text box at the bottom.

\subsection{Data Correction}
After all data has been annotated, we undertake a comprehensive three-stage data correction process to ensure the dataset is of high quality and can be reliably used for training our models. Each stage is designed to address specific issues that may arise during the annotation process.

\textit{1.Coarse Filtering:}
In the first step, we perform a broad initial correction by removing data that contains obvious annotation errors or responses that are irrelevant to the evaluation criteria. This includes any annotations where the feedback from the annotators does not align with the predefined evaluation dimensions or scoring guidelines. Additionally, we remove samples where the video or text prompt is unclear or too ambiguous.

\textit{2.Iterative Refinement:}
Once the coarse filtering step is completed, we split the dataset into two halves. The first half is used to train an initial version of the reward model. This model is then applied to the second half of the dataset to generate annotations for those samples. We compare the model-generated annotations with the original human annotations to assess their alignment. If the model's annotations closely match the human-provided ratings and rationales, they are retained. However, in cases of significant disagreement, the conflicting samples are flagged for further review by human annotators. These annotators manually decide which annotations are correct and should be retained, based on the evaluation criteria.
After correcting and validating the second half of the data, it is used to retrain the reward model. The retrained model is then re-applied to the first half of the dataset to improve the quality of its annotations. 

\textit{3.Final Integration:}
In the final step of the correction process, we use the fully corrected dataset to train a final version of the reward model. This model is then applied to re-annotate any data that was removed in the first stage during the coarse filtering step. We review the newly annotated data and assess whether the model's predictions are consistent with human judgment. The annotations that pass this review are re-integrated into the dataset.

Through these rigorous steps, we ensure that our data correction process effectively identifies and rectifies any inconsistencies, ambiguities, or errors in the dataset.

\begin{table}[t]
\setlength
\tabcolsep{6pt}
\centering
\small
\caption{\textbf{Evaluation results on VideoScore benchmark}. We assess baselines across visual quality (VQ), temporal consistency (TC), dynamic degree (DD), text-to-video alignment (TA), and factual consistency (FC) and compute the average accuracy.}

\begin{tabular}{ccccccc}

		\toprule
     \multirow{2}{*}{Method} &    \multicolumn{5}{c}{\textbf{VideoScore Benchmark}} \\
          & VQ &
	\makecell[c]{TC}
      & \makecell[c]{DD}
  &\makecell[c]{TA} &\makecell[c]{FC}

       \\

       		\midrule
          
    CogVideoX-2B  &  2.86   &   2.78      &  2.65    &  2.91 &   2.71  \\
   
    CogVideoX-5B  &  3.01   &    2.94     &   2.73   & 2.96  & 2.87\\

		\midrule

 		\multirow{1}{*}{CogVideoX-2B-LiFT}   & \textbf{3.18} &    \textbf{2.99}  &  \textbf{2.96}    &  \textbf{3.04} & \textbf{ 2.90 }  \\
		\bottomrule

	\end{tabular} \\

    \vspace{-0.5cm}
\label{tab:videoscore}
\end{table}

\section{More Experiments}
\subsection{Implementation Details}
We train \ourmethod-13B with batch size of 16 for 20 epochs, using LoRA \cite{hu2021lora} rank of 128 and alpha value of 256. For \ourmethod-40B, we used batch size of 8 for 20 epochs, with LoRA rank of 64 and alpha value of 128. 
For the vision towers, we employ \textit{siglip-so400m-patch14-384} \cite{zhai2023sigmoid} for \ourmethod-13B and \textit{InternViT-6B-448px-V1-2} \cite{chen2024internvl} for \ourmethod-40B. Both vision towers, along with their MLP projectors, are frozen during training, and only the linear layers of the LLM are fine-tuned.
Optimization is performed using the AdamW optimizer with a base learning rate of 1e$^{-5}$. A cosine learning rate scheduler is employed, incorporating a warmup ratio of 3e$^{-2}$. The training process is conducted on 8 NVIDIA H100 GPUs. The visualization of training loss is shown in Fig. \ref{fig:loss_plot}. 
\begin{table*}[!ht]
\setlength\tabcolsep{4pt}
\centering
\small
\caption{\textbf{Quantitative results on video assessment metrics}. The first seven metrics correspond to the \textit{Quality} type, while the remaining correspond to the \textit{Semantic} type. ``RM'' denotes the ``Reward Model''.}

\begin{tabular}{ccccccccccc}

		\toprule
         Models & \makecell[c]{RM \\ Size} &
	\makecell[c]{Subject\\ Consistency}
      & \makecell[c]{Background\\ Consistency}
  &\makecell[c]{Aesthetic \\Quality} &\makecell[c]{Imaging \\Quality}

&\makecell[c]{Temporal \\Flickering}
&\makecell[c]{Motion \\Smoothness}
& \makecell[c]{Dynamic \\Degree}
& \makecell[c]{Human \\Action}

       \\
		\midrule

 		T2V-Turbo  &     &  95.6       & 97.01     & 63.2  &73.36 & 97.19 &97.16 &56.67 &95.00    \\
		\midrule
                
          \multirow{1}{*}T2V-Turbo-LiFT  &         40B     & \textbf{96.5}             & \textbf{97.41}    &\textbf{63.88}   &\textbf{73.91} &\textbf{97.53}& \textbf{98.01} & \textbf{68.44} & \textbf{96.00}   \\
		\bottomrule

        \toprule
                
		Models  & \makecell[c]{RM \\ Size} &
		\makecell[c]{Color}
       & \makecell[c]{Spatial \\Relationship}
  &\makecell[c]{Scene} &\makecell[c]{Temporal \\Style}
&\makecell[c]{Overall \\Consistency}
% &\makecell[c]{Temporal \\Flickering}
% &\makecell[c]{Motion \\Smoothness}
% & \makecell[c]{Dynamic \\Degree}
&\makecell[c]{Object \\Class}
&\makecell[c]{Multiple \\Objects}
&\makecell[c]{Appearance \\Style}
       \\
		\midrule

		T2V-Turbo &     &  90.09     & 36.36     & 55.09  &25.51 & 28.17 &95.63 &48.86 &24.40      \\
		\midrule
                
            \multirow{1}{*}T2V-Turbo-LiFT   
        
               &  40B    & \textbf{92.63}             & \textbf{44.81}    &\textbf{58.71}   &\textbf{26.03} &\textbf{30.22}& \textbf{96.12} & \textbf{56.81} & \textbf{26.17}   \\
		\bottomrule

	\end{tabular} \\
\vspace{-0.3cm}

\label{tab:turbo_vbench_compare}
\end{table*}

% \mypara{Evaluation Metrics.} 
we employ Vbench \cite{Huang2023VBenchCB}, a comprehensive benchmark suite to assess the performance of T2V generation, which decomposes ``video generation quality'' into specific, hierarchical, and disentangled dimensions. Each dimension is evaluated using tailored prompts and specialized evaluation methods. 
The evaluation prompts are optimized using Qwen2.5-72B-Instruct \cite{yang2024qwen2} since the CogVideoX \cite{yang2024cogvideox} model is trained with long prompts.

% \subsection{Evaluation on GenAI-Bench}
% We compare \ourmethod with the video assessment model VideoScore \cite{he2024videoscore} on GenAI-Bench \cite{jiang2024genai} as shown in Tab. \ref{tab:genAI}. Since our approach evaluates each video individually using only three discrete categories (``Good'', ``Bad'', ``Normal''), it naturally results in a higher occurrence of ties. This makes it challenging to directly compare performance when ties are included. However, after removing ties, \ourmethod demonstrates a clear advantage over the baseline methods, suggesting that it effectively differentiates video quality and provides more reliable assessments.

\subsection{Evaluation on VideoScore Benchmark}
We compare the performance of CogVideoX-2B, its fine-tuned version CogVideoX-2B-LiFT, and CogVideoX-5B using VideoScore \cite{he2024videoscore} as the assessment model. The results in Tab. \ref{tab:videoscore} further demonstrate the effectiveness of our \ourpipeline.

\subsection{Qualitative Results of \ourmethod}
To demonstrate the effectiveness of our \ourmethod in providing nuanced evaluations that align with human preferences, we provide several qualitative results in Fig. \ref{fig:critic_case1} and Fig. \ref{fig:critic_case2}.

\subsection{Reward-weighted Learning on T2V-Turbo}
To demonstrate the robustness of our proposed pipeline, we apply it to T2V-Turbo \cite{li2024t2v}. The results are presented in Tab. \ref{tab:turbo_vbench_compare}.

\subsection{More Qualitative Comparison}
We provide more qualitative comparison in Fig. \ref{fig:more_compare1} and Fig. \ref{fig:more_compare2}.

\subsection{Human Evaluation}
We conduct a human evaluation to compare the performance of CogVideoX-2B, CogVideoX-5B, and the fine-tuned version CogVideoX-2B-LiFT. To facilitate this process, we design a web-based user interface as illustrated in Fig. \ref{fig:human_ui}. 

% Specifically, for each method, we generate a set of videos using the same 120 text prompts to ensure consistency in the evaluation.
% Next, we perform pairwise comparisons of our method against CogVideoX-2B and CogVideoX-5B, respectively. For each comparison, human raters are asked to determine which method performs better or to indicate a tie (i.e., both methods produce similar results) based on three evaluation criteria: semantic consistency, motion smoothness, and video fidelity. Each query is independently assessed by six human raters.

% The results are summarized in Fig. \ref{fig:human_evaluation}, where we report the percentage of queries receiving positive votes for each method. Additionally, we highlight the percentage of queries achieving a majority consensus (at least three out of six positive votes) in the black box. These results underscore the advantages of our fine-tuned model, CogVideoX-2B-LiFT, over its counterparts.

We also utilize our reward model, \ourmethod, to compare the videos generated by the three methods. The evaluation results are presented in Fig. \ref{fig:reward_compare}, which demonstrate that the rankings produced by our reward model are highly consistent with the results of human evaluations across all three dimensions. This alignment further validates that our reward model effectively captures human preferences, providing reliable and interpretable feedback for T2V model alignment.

\subsection{Ablation Study of Real Video-Text Dataset}
During our T2V model alignment, we incorporate a real video-text dataset, $\mathcal{D}^{\mathrm{real}}$, to address the limitations of relying solely on synthesized datasets: synthesized videos often exhibit low temporal consistency, which hinders the model’s ability to maintain coherent subject alignment across frames and generate smooth motion dynamics.

\begin{figure}[t]

    \centering
    \includegraphics[width=1\linewidth]{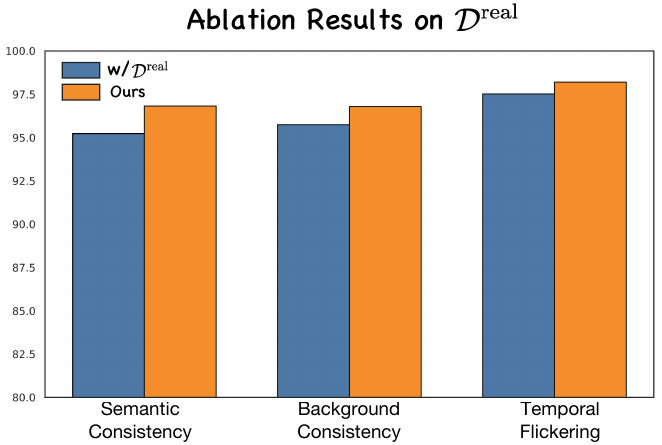}
    \caption{\textbf{Visualized ablation results on $D^{real}$}. We compare the performance of our CogVideoX-2B-LiFT model with and without the inclusion of $D^{real}$ during T2V model alignment.}
    \label{fig:ablation_real}

\end{figure}

\begin{figure*}[!thb]

    \centering
    \includegraphics[width=1\linewidth]{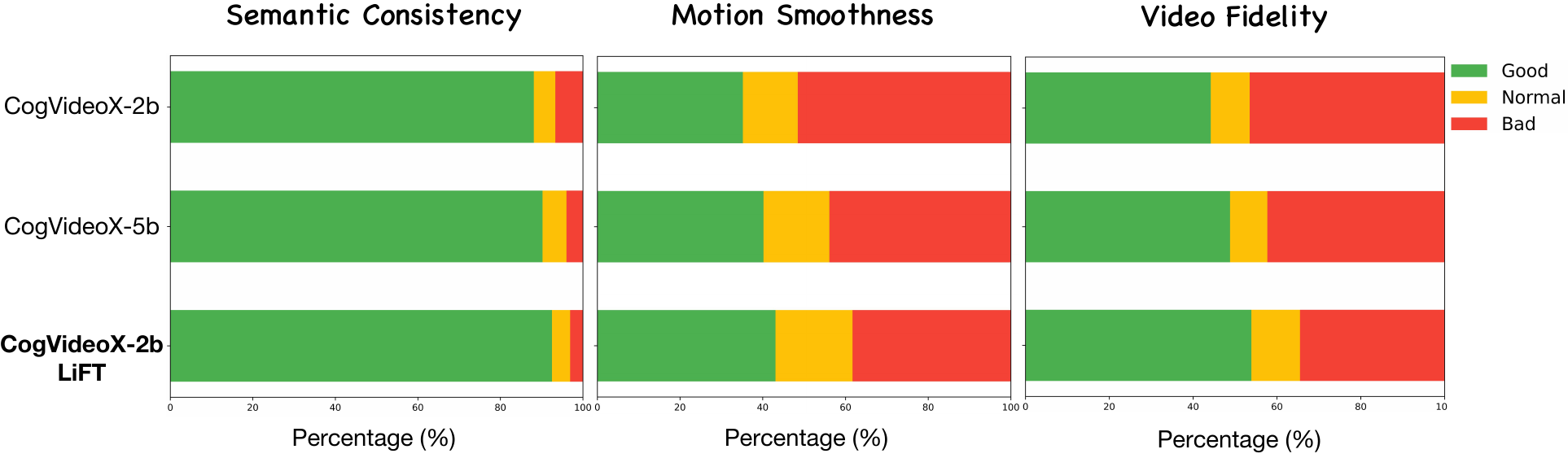}
    \caption{\textbf{Visualized results of comparison evaluated by \ourmethod}.
We leverage \ourmethod to evaluate and compare the performance of CogVideoX-2B, CogVideoX-5B, and our proposed CogVideoX-2B-LiFT.}
    \label{fig:reward_compare}

\end{figure*}

\begin{figure*}[t]

    \centering
    \includegraphics[width=0.9\linewidth]{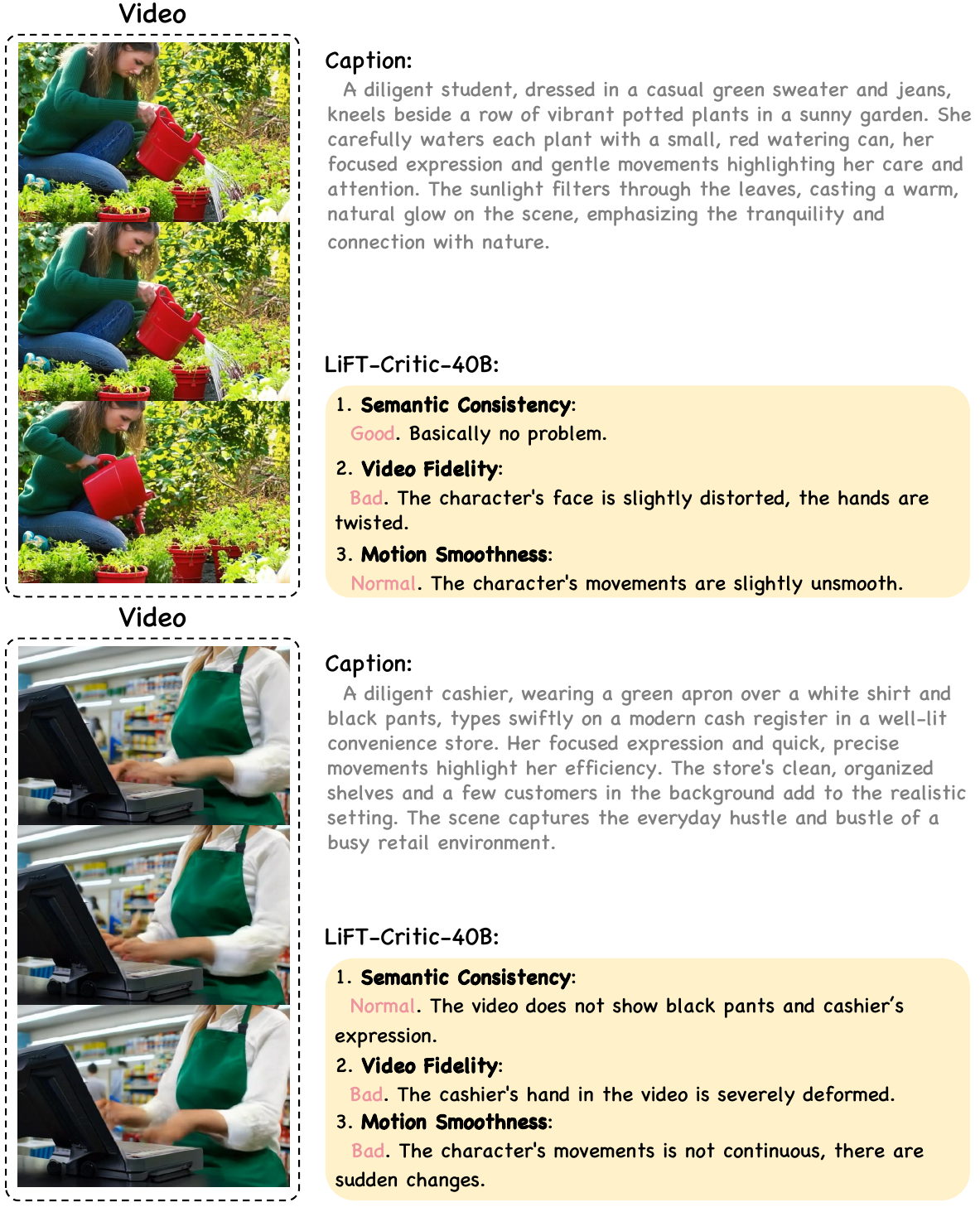}
    \caption{\textbf{Qualitative results of \ourmethod}. We present several case studies illustrating how our \ourmethod evaluates synthesized videos.}
    \label{fig:critic_case1}
    
\end{figure*}

\begin{figure*}[t]

    \centering
    \includegraphics[width=0.9\linewidth]{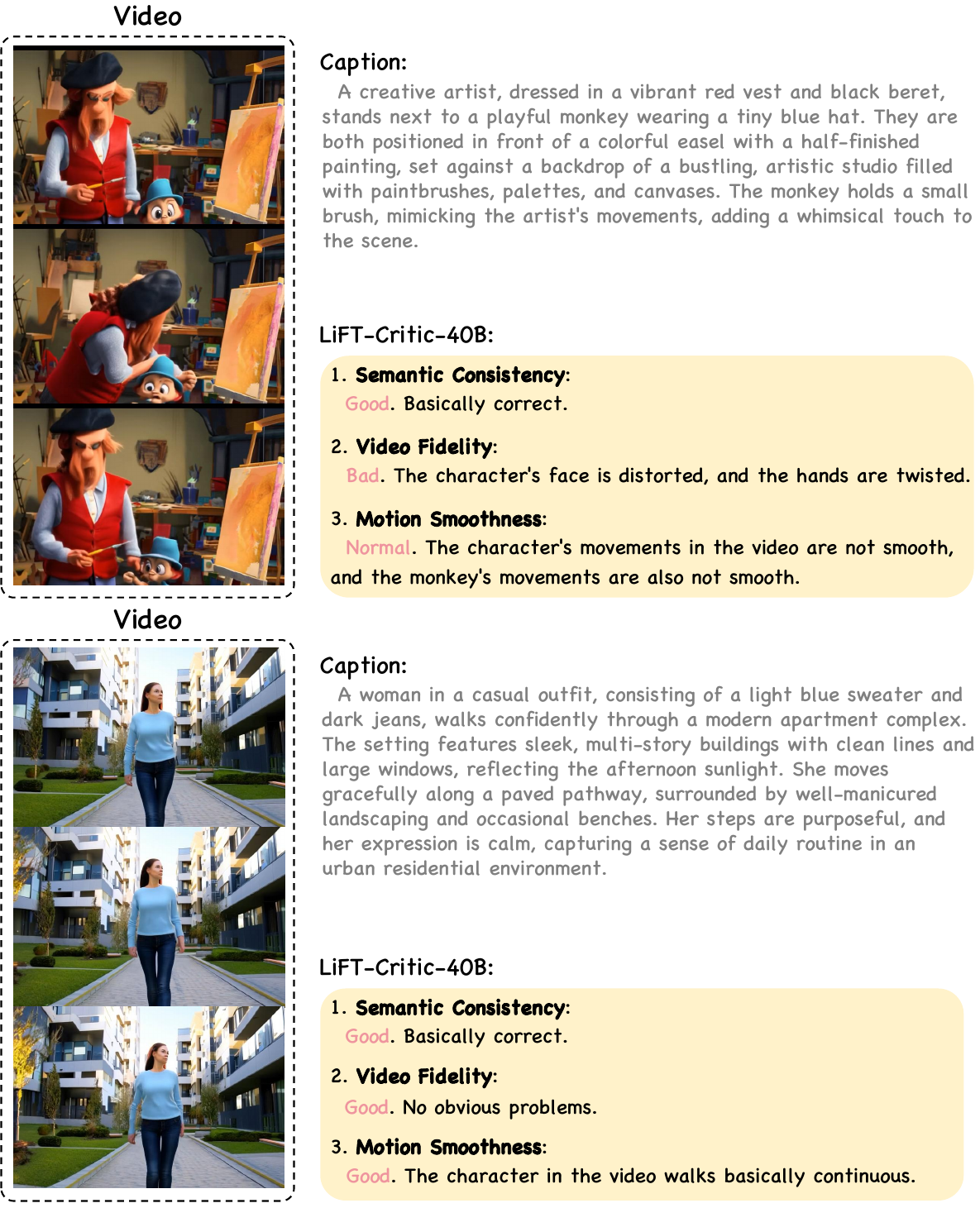}
    \caption{\textbf{Qualitative results of \ourmethod}. We present several case studies illustrating how our \ourmethod evaluates synthesized videos.}
    \label{fig:critic_case2}
    
\end{figure*}

\begin{figure*}[t]

    \centering
    \includegraphics[width=0.8\linewidth]{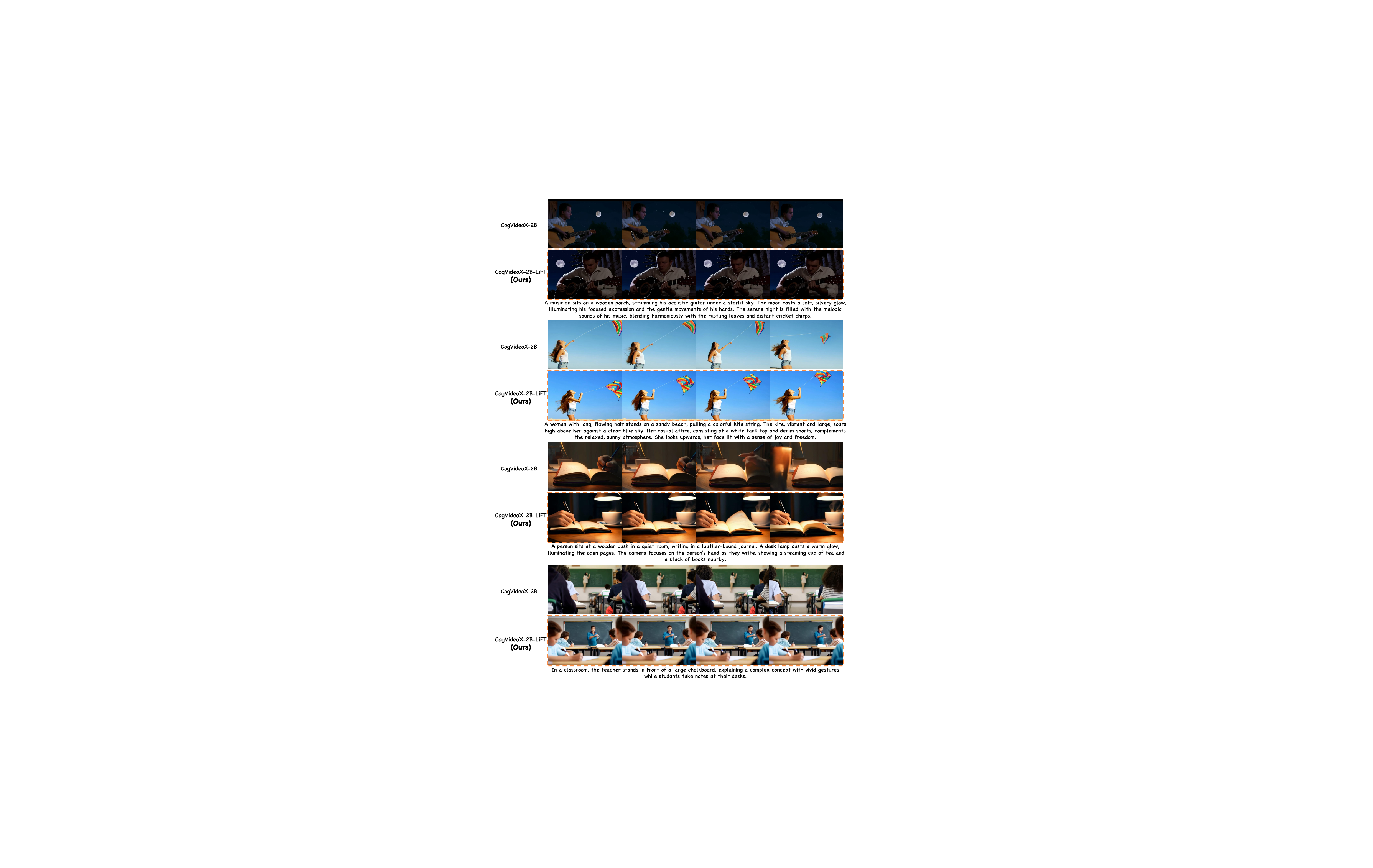}
    \caption{\textbf{More qualitative comparison}. We compare the performance of CogVideo-2B and its LiFT fine-tuned version CogVideo-2B-LiFT.}
    \label{fig:more_compare1}
    \vspace{-0.1cm}
\end{figure*}

\begin{figure*}[t]

    \centering
    \includegraphics[width=1\linewidth]{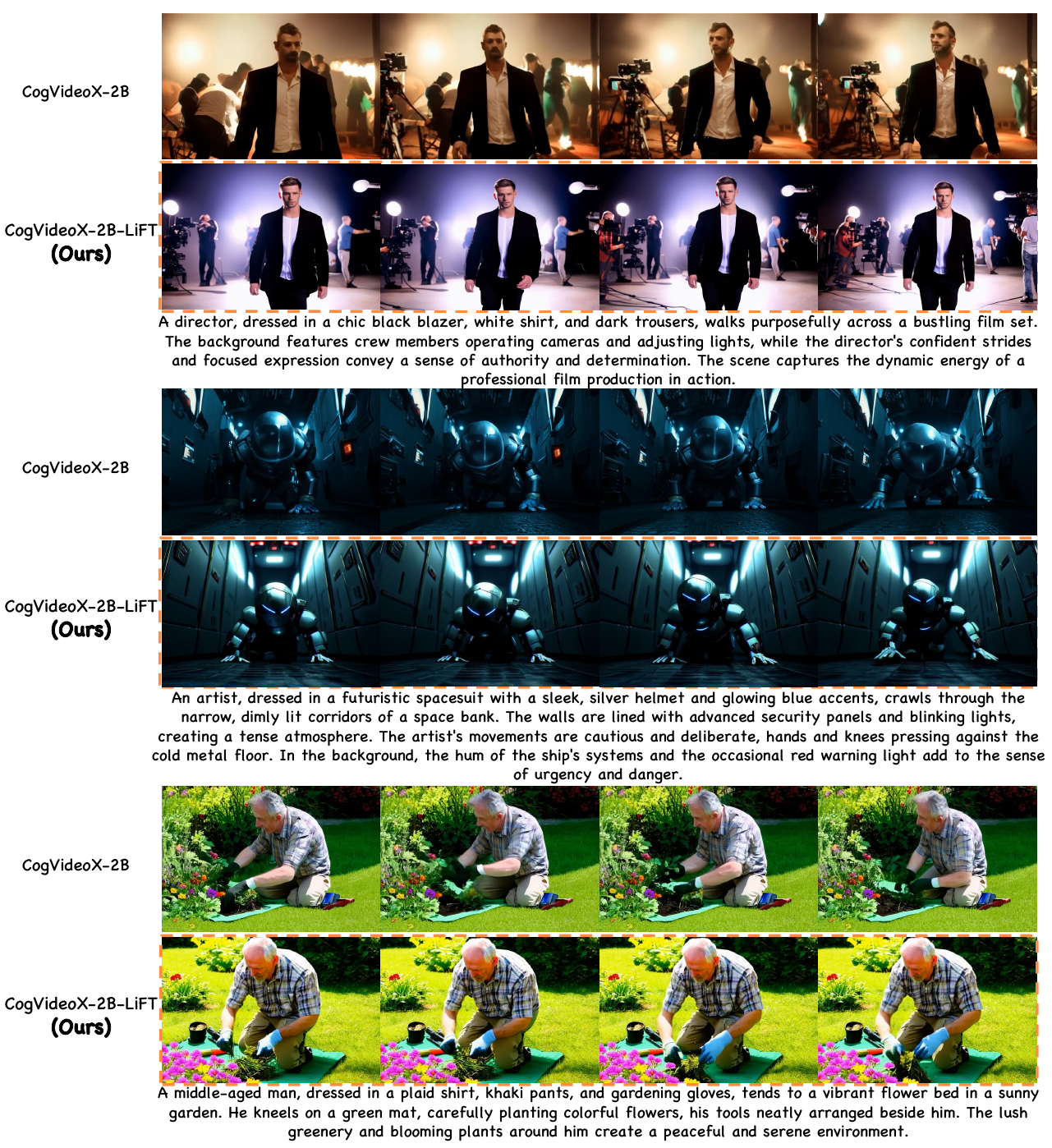}
    \caption{\textbf{More qualitative comparison}. We compare the performance of CogVideo-2B and its LiFT fine-tuned version CogVideo-2B-LiFT.}
    \label{fig:more_compare2}
    \vspace{-0.1cm}
\end{figure*}

\begin{figure*}[t]

    \centering
    \includegraphics[width=1\linewidth]{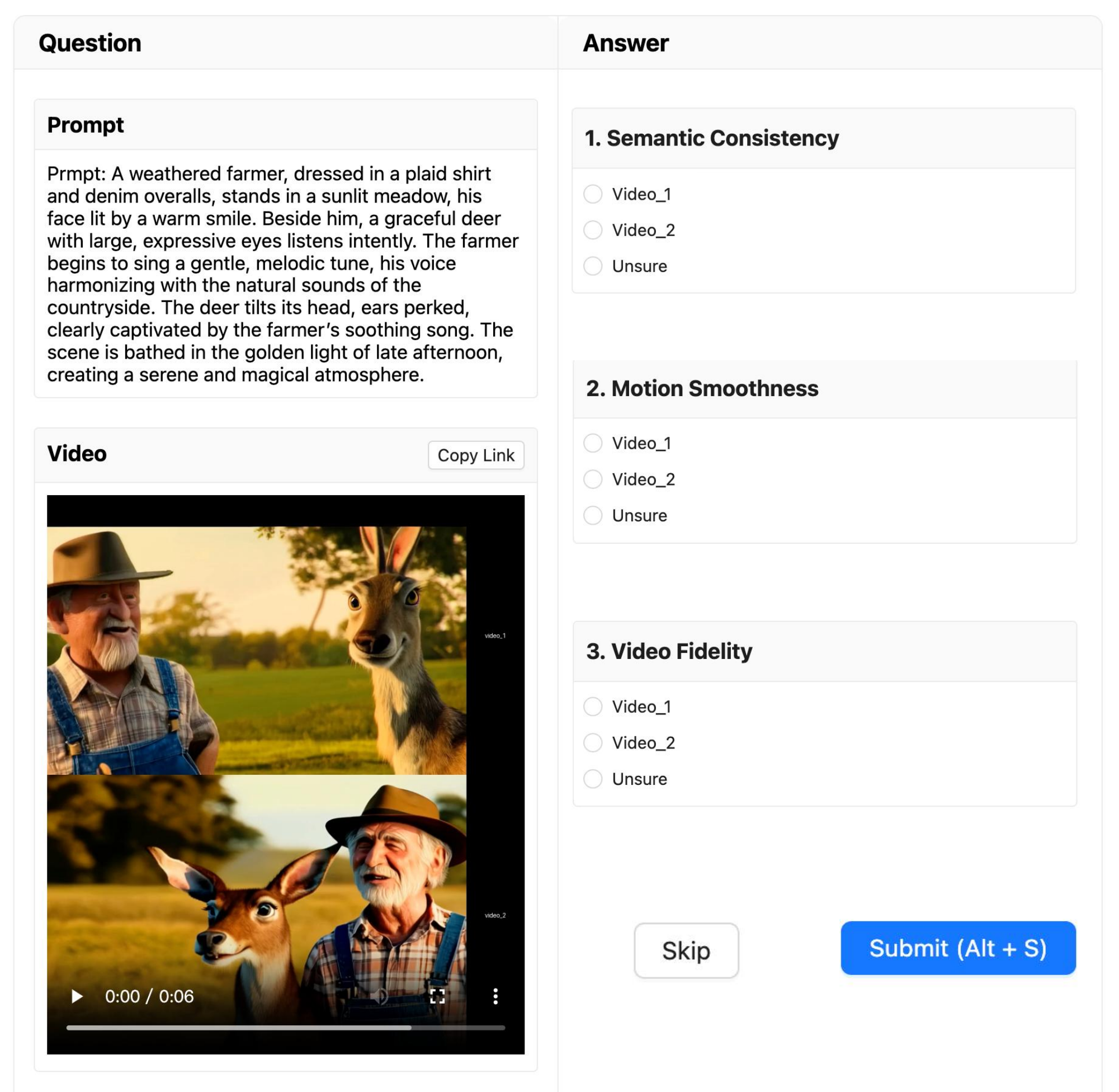}
    \caption{\textbf{An illustration of Human evaluation UI}. We design a web-based user interface to streamline the human evaluation process.}
    \label{fig:human_ui}

\end{figure*}

To evaluate its effectiveness, we conduct an ablation study, with results presented in Fig. \ref{fig:ablation_real}. Specifically, we compare the performance of our full CogVideoX-2B-LiFT model to an ablated version (excluding $\mathcal{D}^{\mathrm{real}}$) using VBench's \cite{Huang2023VBenchCB} temporal quality-related metrics: subject consistency, background consistency, and temporal flickering. The ablated model exhibits significant declines across all three metrics, underscoring the critical role of real data in enhancing temporal quality.
These results demonstrate that incorporating $\mathcal{D}^{\mathrm{real}}$ effectively grounds the model in realistic frame-to-frame dynamics, enabling it to produce videos with superior semantic and temporal fidelity. This aligns closely with our design objectives, reaffirming the importance of blending synthesized and real-world data to achieve robust performance.

\section{Societal Impacts}
Our work aims to improve the alignment of text-to-video generation models with human preferences, enhancing their usability in applications such as educational content creation, video summarization, and creative media production.
% By incorporating human feedback that includes both ratings and rationales, we address the interpretability gap in existing evaluation frameworks, fostering more reliable and user-aligned video generation systems.

However, our method also introduces potential societal risks. First, as the reward model depends on human annotations, biases present in the data may propagate into the model, potentially leading to unintended or unfair outcomes in specific applications. Therefore, the enhanced alignment with human preferences may inadvertently amplify harmful content if the underlying feedback data contains inappropriate or unethical biases. We have taken steps to mitigate these risks by employing diverse and carefully curated datasets and emphasizing transparency in the annotation and training processes.

In summary, while our work primarily targets technical advancements, its broader societal implications require continuous scrutiny to ensure ethical and equitable deployment. We encourage the community to further evaluate and refine such systems to maximize their positive impact while minimizing potential harm.

\section{Limitations and Future Works}
Our work has several limitations and opens avenues for future exploration:

\mypara{Expanding the diversity and scope of the dataset}. Currently, our curated dataset only focuses on three primary objective dimensions, i.e., video fidelity, motion smoothness, and semantic consistency. Additionally, it is limited to a specific set of text categories (i.e., human, animal, place, action) and relies on a relatively simple feedback score (good, normal, bad). This constrained scope reduces the diversity of human feedback data. Therefore, expanding the dataset to include more evaluation dimensions, subjective text categories (e.g., artistic style), and richer feedback formats, such as ranking or multi-level annotations, represents a valuable direction for future research.

\mypara{Exploring alternative optimization strategies.} 
Our work employs reward-weighted likelihood maximization to align the T2V model. However, integrating reinforcement learning from human feedback (RLHF), as demonstrated in language models \cite{ouyang2022training}, may offer further advantages. RLHF allows for online sample generation during training and incorporates KL-regularization, potentially mitigating overfitting to specific reward functions and fostering a more balanced improvement across evaluation metrics.

\section{Ethical Statement}
In this work, we affirm our commitment to ethical research practices and responsible innovation. To the best of our knowledge, this study does not involve any data, methodologies, or applications that raise ethical concerns. All experiments and analyses were conducted in compliance with established ethical guidelines, ensuring the integrity and transparency of our research process.

\begin{table*}[t]
\centering
\caption{\textbf{Categories and examples.} This table displays a subset of our category list used to construct prompts.}

\begin{tabular}{|c|p{12cm}|}
\hline 
\textbf{Category} & \multicolumn{1}{c|}{\textbf{Examples}} \\
\hline
\textbf{Human} & man, woman, children, people, student, teacher, worker, doctor, nurse, engineer, lawyer, police officer, firefighter, chef, waiter, cashier, janitor, farmer, artist... \\
\hline
\textbf{Land Anim.} & horse, antelope, chicken, dog, cat, ch, badger, bat, bear, beaver, bison, bobcat, buffalo, camel, capybara, cheetah, chimpanzee, cobra, coyote, deer, dingo, elephant, mammoth, ferret, fox, gazelle, gerbil, giraffe, gorilla, hamster, hare, hedgehog, hippopotamus, hyena, ibex, jackal, jaguar, kangaroo, koala, leopard, lion, lizard, mammoth, meerkat, mongoose, monkey, moose, mouse, otter, panda, panther... \\
\hline
\textbf{Other Anim.} & Fish, Squid, Octopus, Shark, Dolphin, Whale, Turtle, Seahorse, Starfish, Crab, Lobster, Snail, Clam, Oyster, Jellyfish, Anemone, Ray, Eel, Cuttlefish, Manta Ray, Pufferfish, Angelfish, Butterfly Fish, Parrotfish, Grouper, Mackerel, Sardine, Herring, Cod, Salmon, Trout, Bass, Halibut, Flounder, Tuna, Marlin, Swordfish, Nudibranch, Sea Cucumber, Sea Urchin, Sand Dollar, Zebrafish, Pipefish, Mole Crab, Ghost Shrimp, Hermit Crab, Coral, Bird, Eagle... \\
\hline
\textbf{Place} & mountains, forests, parks, city streets, Tokyo streets, Shanghai streets, beaches, rivers, lakes, deserts, oceans, valleys, canyons, waterfalls, cliffs, glaciers, volcanoes, geysers, hot springs, caves, swamps, wetlands, tundra, grasslands, savannas, rainforests, deciduous forests, boreal forests, taiga, orchards, vineyards, farmlands, meadows, heaths, woodlands, thickets, scrublands, chaparral, tropical dry forests, temperate rainforests, temperate deciduous forests, temperate grasslands, temperate shrublands, subtropical deserts, subtropical dry forests, subtropical moist forests, subtropical shrublands, subtropical wet forests, subtropical woodlands, urban parks, botanical gardens, zoological parks, amusement parks, historical sites, museums, galleries, stadiums, arenas, theaters, cinemas, restaurants, shops, markets, supermarkets, malls, hotels, resorts, airports, train stations, bus stops, highways, bridges... \\
\hline
\textbf{Imagine Place} & car in a pot, airplane in a bowl, train in the cloud, elephant on a spoon, guitar in a teacup, house on a leaf, shoe in a tree, phone in a puddle, piano in a bubble, dinosaur in a bathtub, moon in a saucer, book in a galaxy, clock in a flower, computer in a handbag, lamp in a well, camera in a fridge, bicycle in a cake, hat in a pool, sun in a shoebox, camera in a birdcage, laptop in a fishbowl, glasses in a garden, bed in a suitcase, mirror in a pond, television in a tent, comb in a forest, microwave in a hat, towel in a jar, oven in a backpack, table in a bubble, chair in a cloud, toothbrush in a lake, soap in a park, perfume in a cave, toilet in a spaceship, key in a painting, window in a book, door in a cake, stove in a suitcase, sink in a meadow... \\
\hline
\textbf{Easy Act.} & running, walking, jumping, crawling, wiping, caressing, swimming, dancing, singing, drawing, writing, reading, cooking, eating, drinking, climbing, hiking, fishing, sleeping, sitting, standing, kicking, throwing, catching, pushing, pulling, lifting, digging, painting, typing, listening, talking, laughing, crying, smiling... \\
\hline
\textbf{Difficult Act.} & blinking, breathing, yawning, stretching, sneezing, coughing, scratching, cleaning the car, washing dishes, stir frying, doing laundry, vacuuming the house, mopping the floor, dusting furniture, organizing shelves, ironing clothes, polishing silverware, gardening, watering plants, cooking dinner, baking a cake, making coffee, loading the dishwasher, unloading the dishwasher, taking out the trash, sweeping the porch, raking leaves, shoveling snow, painting a room, fixing a leaky faucet, changing a light bulb, replacing batteries, checking smoke detectors, cleaning windows, scrubbing the bathroom, rearranging furniture, hanging pictures, repairing a fence, washing the dog, feeding pets, walking the dog, cleaning the fish tank, organizing the garage, cleaning the gutters, repairing a bike...\\
\hline

\end{tabular}
\label{tab:category}
\end{table*}

\end{document}